\documentclass[10pt]{article}
\PassOptionsToPackage{table}{xcolor}
\usepackage{paperstyle} 
\usepackage{booktabs,multirow}
\usepackage{amsmath,amssymb}
\usepackage{graphicx}
\usepackage{xcolor}
\usepackage{colortbl,tabularx}
\usepackage{booktabs,multirow,amssymb}
\usepackage{tcolorbox}
\usepackage[normalem]{ulem}
\tcbuselibrary{skins,breakable}

\usepackage{tcolorbox}
\tcbuselibrary{skins}
\usepackage{xcolor}
\usepackage{tcolorbox}
\tcbuselibrary{skins}

\definecolor{TakeawayPaper}{HTML}{F7F3EB}
\definecolor{TakeawayWine}{HTML}{763E35}

\providecommand{\takeaway}[2][1]{}
\renewcommand{\takeaway}[2][1]{%
  \begin{tcolorbox}[
    enhanced,
    colback=TakeawayPaper,
    colframe=TakeawayPaper,
    boxrule=0pt,
    sharp corners,
    boxsep=0pt,
    left=8pt,right=8pt,
    top=5pt,bottom=5pt,
    before skip=6pt,after skip=6pt,
    fontupper=\normalsize\rmfamily\bfseries
  ]
    {\color{TakeawayWine}\itshape Takeaway~#1.}\enspace #2
  \end{tcolorbox}%
}
\usepackage{placeins} 
\usepackage{eso-pic}
\AddToShipoutPictureFG*{%
  \put(\LenToUnit{1.18in},736){\makebox[\textwidth][s]{%
    \includegraphics[height=26pt]{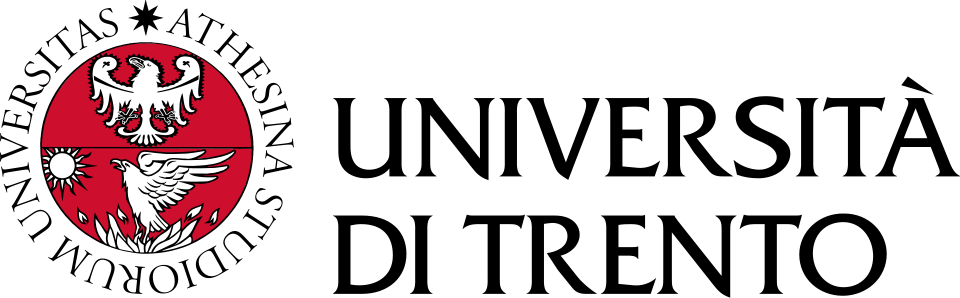}\hfill
    \includegraphics[height=26pt]{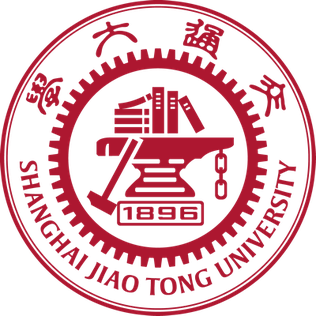}\hfill
    \includegraphics[height=26pt]{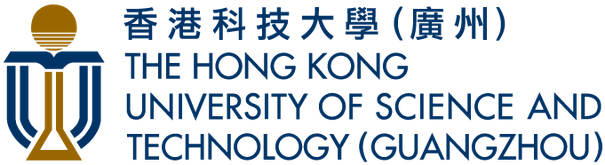}%
  }}%
}
\title{LIFE-VLR: Revisiting Formation, Evolution, and Readout in Visual Latent Reasoning}
\author{\raggedright \mbox{Hangui Lin$^{1,*}$}, \mbox{Yan Shu$^{2,*}$}, \mbox{Zequn Yang$^{1,3,*}$}, \mbox{Xuanming Shang$^{1,3}$}, \mbox{Yuhang Han$^{1,3,4}$}, \mbox{Wenbin Duan$^{5}$}, \mbox{Teng Long$^{2}$}\\[3pt]
{\normalfont \mbox{$^{1}$Shadoweave}\quad
\mbox{$^{2}$University of Trento}\quad
\mbox{$^{3}$Shanghai Jiao Tong University}\quad
\mbox{$^{4}$The Hong Kong University of Science and Technology (Guangzhou)}\quad
\mbox{$^{5}$University of International Relations}}\\[2pt]
}

\begin{document}
\maketitle
\raggedbottom
\begin{abstract}
Visual latent reasoning uses continuous hidden states as intermediate representations and has shown promising gains on multimodal tasks. However, existing systems rely on untested assumptions about where latent states should be extracted, whether they need to change across steps, and what information the model can utilize from them.
We systematically examine these assumptions across the three stages of a latent state's lifecycle: \textbf{formation}, \mbox{\textbf{evolution}}, and \textbf{readout}.
Specifically, we find that (1) during\hspace{0.3em}\mbox{\textbf{formation}}, earlier decoder layers retain stronger visual alignment and often outperform the default final layer as state sources; (2) during \textbf{evolution}, cross-step state variation is not required, whereas preserving visual focus improves accuracy even as state diversity decreases; and (3) during \textbf{readout}, task information is retained in latent states yet underutilized for answer prediction, creating a critical bottleneck alongside visual recovery.
Guided by these findings, we propose a unified training procedure that combines visually aligned layer selection, visual subspace grounding, and a staged curriculum for visual and task readout.
Across nine benchmarks, two frameworks (LVR and MONET), and three backbones spanning 3B to 27B parameters, our approach improves accuracy in 51 of 54 comparisons by up to 17.35 percentage points and maintains gains after reinforcement learning.
\par\vspace{6pt}\noindent
{\fontsize{10pt}{12pt}\selectfont \textbf{$^{*}$Equal contribution}\\
\textbf{Code:} \url{https://github.com/Sammy20207109/LIFE-VLR}\\
\textbf{Correspondence:} yan.shu@unitn.it; linhangui@uir.edu.cn}
\end{abstract}

\section{Introduction}
\label{sec:intro}


Visual reasoning challenges multimodal large language models (MLLMs) to ground relevant image evidence~\cite{shu2025semantics} and draw multi-step inferences to answer a given question. Following its success in text, chain-of-thought (CoT) prompting has been extended to multimodal tasks, improving accuracy through intermediate rationales expressed in language~\cite{wei2022chain,shao2024visualcot}. To preserve perceptual details that language struggles to capture, \emph{thinking with images} further interleaves explicit visual observations and manipulations with textual steps~\cite{shao2024visualcot,hu2024visual}. Building on continuous latent reasoning~\cite{hao2024training}, visual latent reasoning internalizes this process by encoding visual evidence and task information jointly within continuous hidden states, offering a promising paradigm for multimodal inference~\cite{li2025latent,yang2025mirage,wang2025monet}.

\begin{figure}[t]
    \centering
\includegraphics[width=0.328\linewidth]{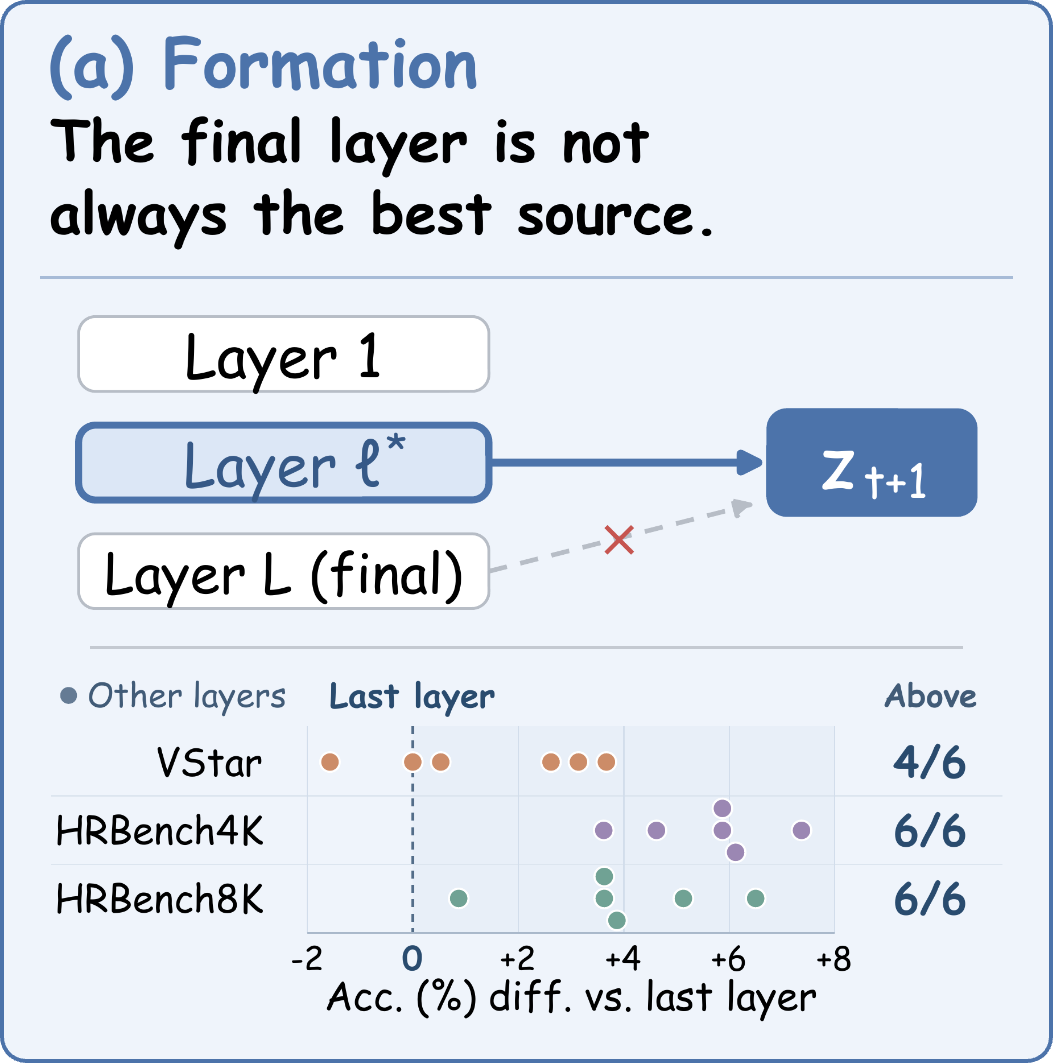}\hspace{3pt}%
\includegraphics[width=0.328\linewidth]{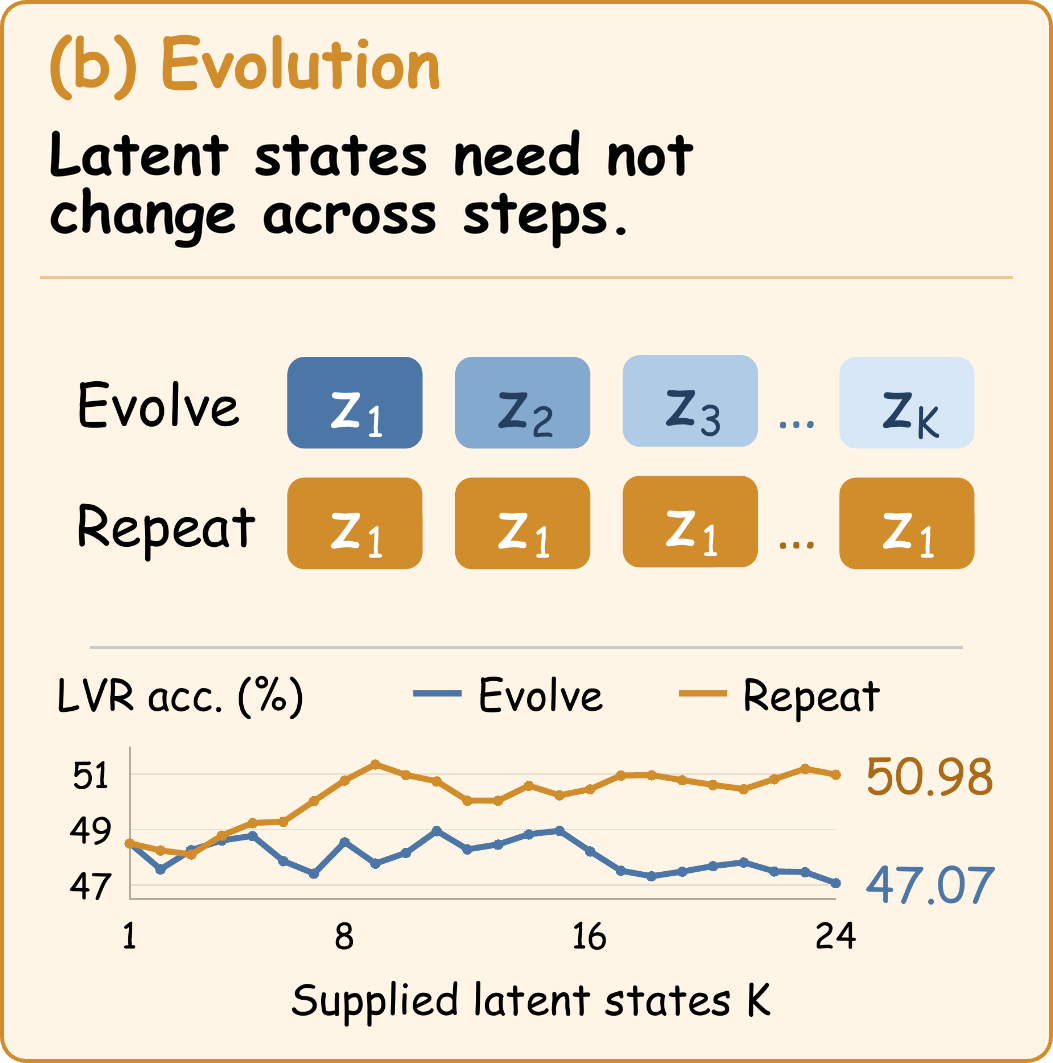}\hspace{3pt}%
\includegraphics[width=0.328\linewidth]{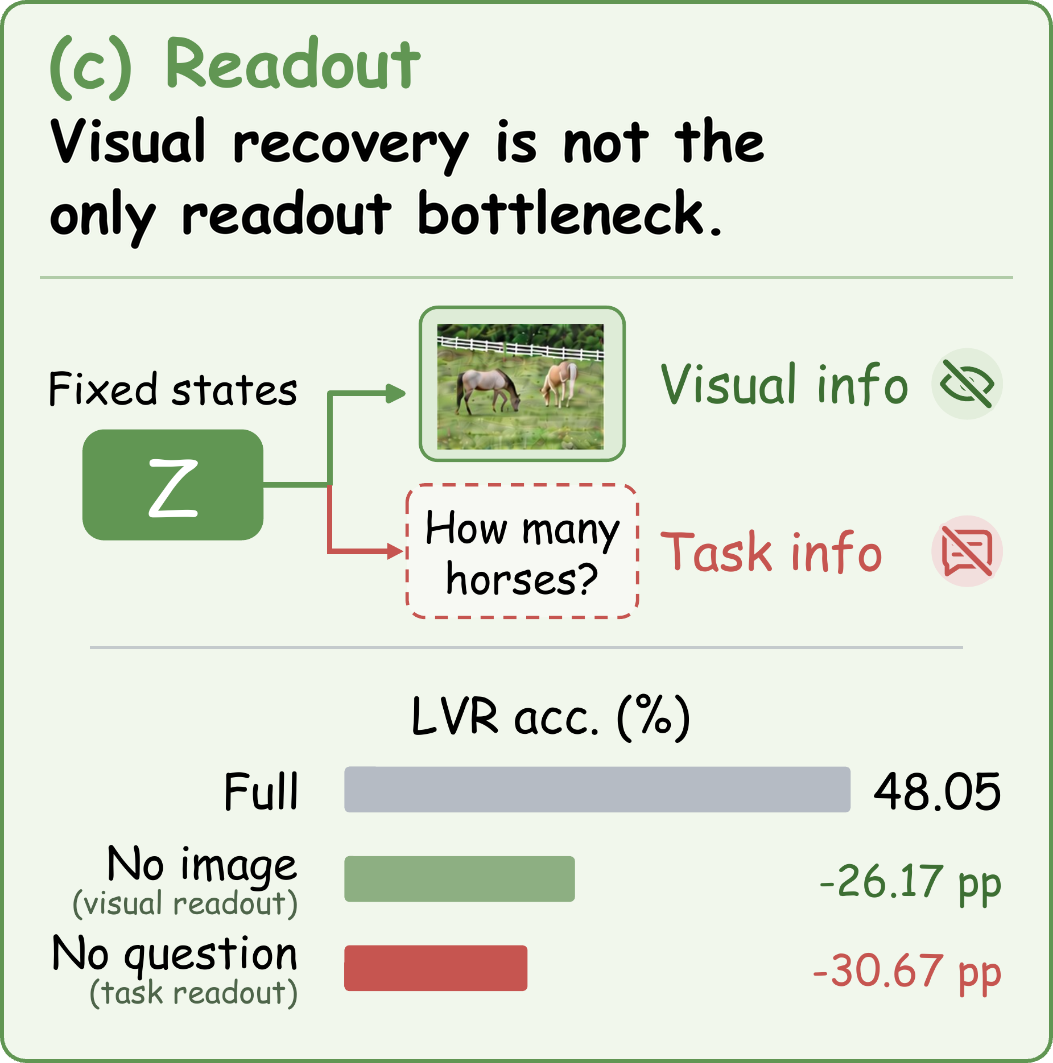}
   \caption{\textbf{Bottlenecks across the lifecycle of visual latent states} (illustrated on LVR-3B). 
\textbf{(a) Formation:} Earlier decoder layers outperform the final layer as latent state sources. 
\textbf{(b) Evolution:} Repeating the initial state $z_1$ outperforms the evolving sequence $z_{1:K}$ at matched lengths $K$.
\textbf{(c) Readout:} Withholding the question degrades accuracy more than withholding the image when reading fixed states $Z$, revealing a critical task-readout bottleneck. 
}
    \vspace{-10pt}
    \label{fig:overview}
\end{figure}


In typical visual latent reasoning frameworks such as LVR~\cite{li2025latent} and MONET~\cite{wang2025monet}, an MLLM processes an image and a question by autoregressively generating a sequence of continuous hidden states, feeding each state back as the next input before decoding the final answer. Throughout this process, existing methods rely on three default practices: they always extract latent states from the final decoder layer, assume that states should progressively vary across steps to advance reasoning, and focus primarily on recovering visual details from these states.
Although recent analyses reveal that latent states often
contribute less than expected to visual
reasoning~\cite{li2026imagination,cui2026ris,zhang2026visual},
whether these default choices underlie the observed limitations
remains unclear. We therefore examine each default at its
corresponding stage in the computational lifecycle of visual
latent states (Fig.~\ref{fig:overview}), namely the choice of
decoder layers for constructing states (\textbf{formation}), the
need for state variation across reasoning steps
(\textbf{evolution}), and the use of visual and task information
carried by the states (\textbf{readout}).

In \textbf{state formation}, existing methods default to the final
decoder layer as the state source, yet earlier layers show stronger
alignment with target image regions and yield better downstream
accuracy, supporting \textbf{extracting latent states before the
final layer}.
In \textbf{state evolution}, whereas recent methods promote
cross-step state diversity~\cite{cui2026ris,li2026prolavit}, simply
repeating the initial latent state can outperform the native
evolving sequence. Native updates weaken the initial state's visual
focus, and steering states toward target visual features improves
accuracy even as diversity decreases, showing that
\textbf{preserving visual evidence across steps can matter more than
maximizing state diversity}.
In \textbf{state readout}, effective answer generation requires
extracting task cues beyond visual details~\cite{zhang2026visual}.
Removing the question during answer prediction causes an even larger
accuracy drop than removing the image, yet adapting readout on the
same cached states partially recovers accuracy. This recovery
shows that \textbf{task-related information is retained in latent
states yet underused by the receiving MLLM}, establishing task
readout as a distinct bottleneck alongside visual readout.

Guided by these findings, we propose \textbf{LIFE-VLR},
a unified training procedure that targets the three identified
bottlenecks without introducing extra model parameters.
Specifically, we extract latent states from visually aligned
decoder layers, combine attention-guided steering with
a region-grounding loss to preserve image evidence across
steps, and strengthen both visual and task readout through
a staged curriculum that withholds the image or question. Our main contributions are as follows:



\begin{itemize}
    \item \textbf{Lifecycle Diagnosis of Visual Latent Reasoning.}
We examine formation, evolution, and readout, revealing three
counter-intuitive bottlenecks: earlier layers show stronger
visual alignment than the final layer, visual focus can matter
more than state diversity, and task information is retained
yet underutilized.

    \item \textbf{A Unified, Bottleneck-Targeted Training Recipe.}
    We translate our diagnostic insights into three complementary
    interventions that improve how latent states are extracted
    across layers, grounded in visual features across steps,
    and decoded for visual and task cues.

    \item \textbf{Broad Gains Across Scales, Frameworks, and RL.}
    Across nine benchmarks, two frameworks (LVR and MONET),
    and three backbones (3B, 8B, and 27B), our approach improves
    accuracy in 51 of 54 comparisons by up to 17.35 percentage
    points, with gains persisting after reinforcement learning.
\end{itemize}


\section{Related Work}
\label{sec:related}


\paragraph{Reasoning with intermediate images.}
Incorporating explicit visual intermediates allows MLLMs to retain and inspect fine-grained visual evidence during multi-step reasoning. One line of work focuses on localizing task-relevant regions through annotated bounding boxes~\cite{shao2024visualcot}, guided visual search~\cite{wu2024vstar}, sequential image crops and zooms~\cite{qi2025cogcom}, or pixel-level grounding masks~\cite{shu2026terrascope}, with recent studies using reinforcement learning to optimize when and where to inspect the image~\cite{zheng2025deepeyes,wang2025pixelreasoner,zhang2025adaptive,lai2026mini}. Beyond region selection, tool-augmented approaches invoke external vision modules via generated programs~\cite{gupta2023visual,suris2023vipergpt}, render auxiliary sketches~\cite{hu2024visual}, edit structured visual inputs~\cite{fu2025refocus}, or execute code for image manipulation~\cite{zhang2025thyme}. A third direction synthesizes new intermediate images as visual subgoals or imagined states~\cite{li2025imagine,chern2025thinking}, supported by large-scale interleaved image-text datasets~\cite{li2025zebra}. While these paradigms rely on explicit regions, external tools, or generated pixels to carry intermediate information, visual latent reasoning performs intermediate computation directly within continuous hidden states.

\paragraph{Visual latent reasoning.}
Visual latent reasoning extends continuous latent reasoning
in language models~\cite{hao2024training,shen2025codi,xu2025softcot}
to multimodal inputs, feeding hidden states that encode visual
evidence back as input embeddings for subsequent reasoning
and answer prediction.
Most methods shape these states with explicit visual targets
by aligning or distilling them toward region and auxiliary-image
features~\cite{li2025latent,yang2025mirage,wang2025monet,jeon2026valr}
or enriching them with attribute-level semantics~\cite{xu2026semantic},
whereas others route visual information through latents without
intermediate supervision~\cite{li2026latent}.
Architectures
also vary in how states interact across steps,
such as injecting selected image embeddings at each
step~\cite{chen2026reasoning}, regularizing cross-step
diversity~\cite{li2026prolavit}, or generating
states
in parallel~\cite{lu2026deeplatent}.
However, recent diagnostic studies question whether these
states are effectively
utilized, showing that
they can often be replaced by noise or boundary markers
with minimal performance loss~\cite{zhu2026leveraging,guo2026beyond}
and respond weakly
to answer-changing image edits~\cite{xiao2026rethinking}.
These studies attribute this underutilization to information
bypassing and representation
mismatches~\cite{li2026imagination,cui2026ris,miao2026gap,viveiros2026holding}
or unexploited latent semantics~\cite{zhang2026visual},
motivating auxiliary objectives that reward latent usage
or add visual supervision~\cite{zhu2026leveraging,xiao2026rethinking}.
Yet these studies focus primarily on
the presence and use of
visual evidence,
leaving default pipeline assumptions intact.
In contrast, we systematically examine the full lifecycle
of latent states across \textbf{formation}, \textbf{evolution},
and \textbf{readout}.

\section{Preliminaries}
\label{sec:preliminaries}

In visual latent reasoning, an MLLM constructs continuous
intermediate states from an image $I$ and a question $Q$,
then uses these states to generate an answer $A$.
Let $Z_t=(z_1,\ldots,z_t)$ denote the latent sequence
after $t$ steps, with $Z_0=\varnothing$ and
$Z=Z_K$ denoting the complete sequence of $K$ states.
At each step, the model forms a new state from the
image, question, and preceding states, then appends it
to the sequence.
For $t=0,\ldots,K-1$, omitting interleaved text
for brevity, we write:
\begingroup
\small
\begin{equation}
\overbrace{
Z_{t+1}=\bigl(
Z_t,\,
\underbrace{
\mathcal{F}\!\bigl(\mathcal{M}_{\theta}(I,Q,Z_t)\bigr)
}_{\text{formation}}
\bigr)
}^{\text{evolution}},
\quad
A\sim
\underbrace{
p_{\theta}(\,\cdot\mid I,Q,Z)
}_{\text{readout}}.
\label{eq:latent_process}
\end{equation}
\endgroup
Here, $\mathcal{M}_{\theta}$ denotes decoder computation
across $L$ layers with parameters $\theta$, $\mathcal{F}$
extracts the new state $z_{t+1}$ from internal layer
representations, and $(Z_t,z_{t+1})$ denotes appending
this state to the existing sequence.

We examine this process from three connected perspectives.
\emph{Formation} concerns how each latent state is
constructed from the image and question, focusing on
which decoder layers are used to form the state.
\emph{Evolution} concerns how the states change as
the sequence develops across steps and whether successive
updates preserve information that supports subsequent reasoning.
\emph{Readout} concerns how the model interprets and uses
the supplied states during answer generation, including
its access to the visual and task-related information
they retain.

\section{Analysis on Visual Latent States}
\label{sec:analysis}
To examine bottlenecks across different latent-reasoning
designs and model scales, we select two representative
systems: LVR~\cite{li2025latent} on Qwen2.5-VL-3B-Instruct,
denoted \textbf{LVR-3B}, and MONET~\cite{wang2025monet}
on Qwen2.5-VL-7B, denoted \textbf{MONET-7B}.
We train LVR-3B following the official LVR SFT procedure
and settings, while MONET-7B uses the officially released
SFT checkpoint.
We then conduct diagnostic evaluations and representation
visualizations on the same 512 image--question pairs
from the GQA validation subset of
Visual-CoT~\cite{shao2024visualcot}.
Sampling and experiment-specific protocols are detailed
in Appendices~\ref{app:v2_analysis_settings}
and~\ref{app:v2_diagnostics}.


\subsection{State Formation}
\label{sec:formation}
\label{sec:intra}
State formation constructs each latent state from decoder
representations, determining the inputs supplied to subsequent
reasoning steps and answer prediction. Conventional frameworks
extract hidden states from the final decoder layer by default,
yet whether this choice outperforms earlier layers remains
untested. To evaluate this default, we test layer sensitivity
during frozen-model inference, compare downstream accuracy
across source layers after full training, and analyze visual
alignment to explain the observed differences.

\begin{figure}
    \centering
    \includegraphics[width=0.8\linewidth]{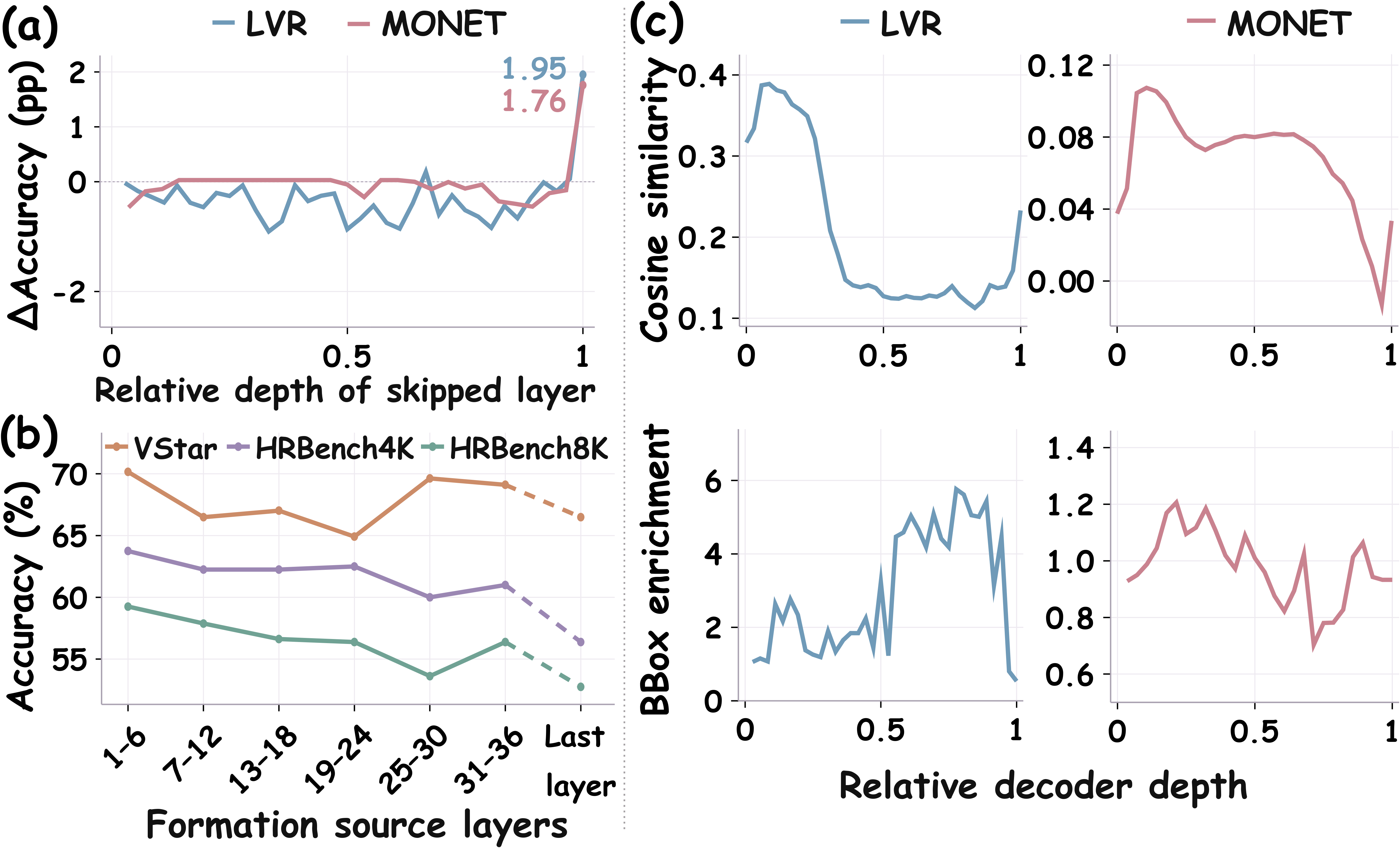}
    \caption{\textbf{State formation across decoder depth.} (a) Accuracy changes when bypassing individual decoder layers during frozen inference. (b) Final benchmark accuracy across sampled source-layer ranges versus the default final layer under matched-budget training. (c) Target-crop cosine similarity (top) and BBox enrichment (bottom) across decoder depth.}
    \label{fig:formation}
    \vspace{-10pt}
\end{figure}

\paragraph{The final layer is suboptimal in both inference and training.}
We first bypass individual decoder layers during latent-state
generation in frozen LVR-3B and MONET-7B checkpoints
(protocols in Appendix~\ref{app:v2_diagnostic_formation}).
Although both models were trained to extract latent states from
the final layer, bypassing that layer at test time raises
accuracy by 1.95 and 1.76 percentage points, respectively,
whereas bypassing earlier layers generally degrades accuracy
(Fig.~\ref{fig:formation}a). To test whether this pattern
persists when weights adapt to alternative sources, we train
LVR-3B under official settings with matched update budgets
across six-layer source ranges (1--6 through 31--36), randomly
sampling a layer within each assigned range during training and
inference. Across \mbox{VStarBench}~\cite{wu2024vstar}, HRBench4K, and
HRBench8K~\cite{hrbench}, these
range-based configurations outperform the fixed final-layer
baseline in 16 of 18 comparisons, with the shallowest range
(1--6) achieving the highest accuracy on every benchmark
(Fig.~\ref{fig:formation}b).

\paragraph{Earlier layers retain stronger visual alignment than the final layer.}
To understand why earlier layers form stronger latent states,
we probe hidden states at latent positions across decoder depth
in both models using two metrics
(Appendix~\ref{app:formation_visual_alignment}):
\emph{target-crop cosine similarity} between each layer's
hidden state and the mean visual embedding of an
independently encoded crop of the annotated target box, and
\emph{BBox enrichment}, measuring attention to image tokens
within that box, normalized by image-directed attention and
the box's share of image tokens.
Both metrics peak well before the final layer
(Fig.~\ref{fig:formation}c). Shallow layers achieve substantially
higher target-crop similarity than the final layer in both
models, despite explicit visual supervision at the output layer
(visible as the small rebound at relative depth $1$). Spatial focus
follows a similar trend, with BBox enrichment peaking in shallow
layers for MONET and middle-to-late layers for LVR. This pattern
is consistent with the final decoder layer adapting
representations toward discrete token prediction, which shifts
hidden states away from visual features and dilutes spatial
focus, motivating visual alignment as a criterion for
source-layer selection.

%

\takeaway{The final-layer default is suboptimal, as earlier layers preserve stronger visual alignment and yield higher downstream accuracy, motivating visually aligned layer selection.}


\subsection{State Evolution}
\label{sec:state_evolution}
\textcolor{black}{State evolution governs how latent states change across steps
as the sequence $Z_K$ develops. Recent methods encourage or
favor cross-step state diversity to advance visual
reasoning~\cite{li2026prolavit,cui2026ris}, yet it remains
unclear whether latent states must vary across steps and what
information successive updates should preserve.
}

\paragraph{\textcolor{black}{Effective latent states need not vary across steps.}}

\textcolor{black}{To isolate cross-step state variation from multi-step
computation, we match the number of latent positions $K$ and
vary only the vectors supplied at those positions. Using frozen
LVR-3B and MONET-7B checkpoints on the same 512 examples, we
cache trajectories from full image--question inputs under their
default latent budgets (eight steps for LVR and ten for MONET;
applied within each latent segment for MONET). Across supplied lengths
up to each budget ($K=1,\ldots,8$ and $K=1,\ldots,10$,
respectively), we compare the native sequence $(z_1,\ldots,z_K)$
against $K$ copies of $z_1$ under identical inputs and decoding
settings, allowing the decoder to continue computing across
positions while keeping the supplied state fixed. To test whether accuracy depends on which state is reused,
we replace $z_1$ with each later cached state $z_t$ ($t>1$)
and repeat it at the default budget (details in
Appendix~\ref{app:evolution_replay}).}

\begin{figure*}[t]
    \centering
    \includegraphics[width=\linewidth]{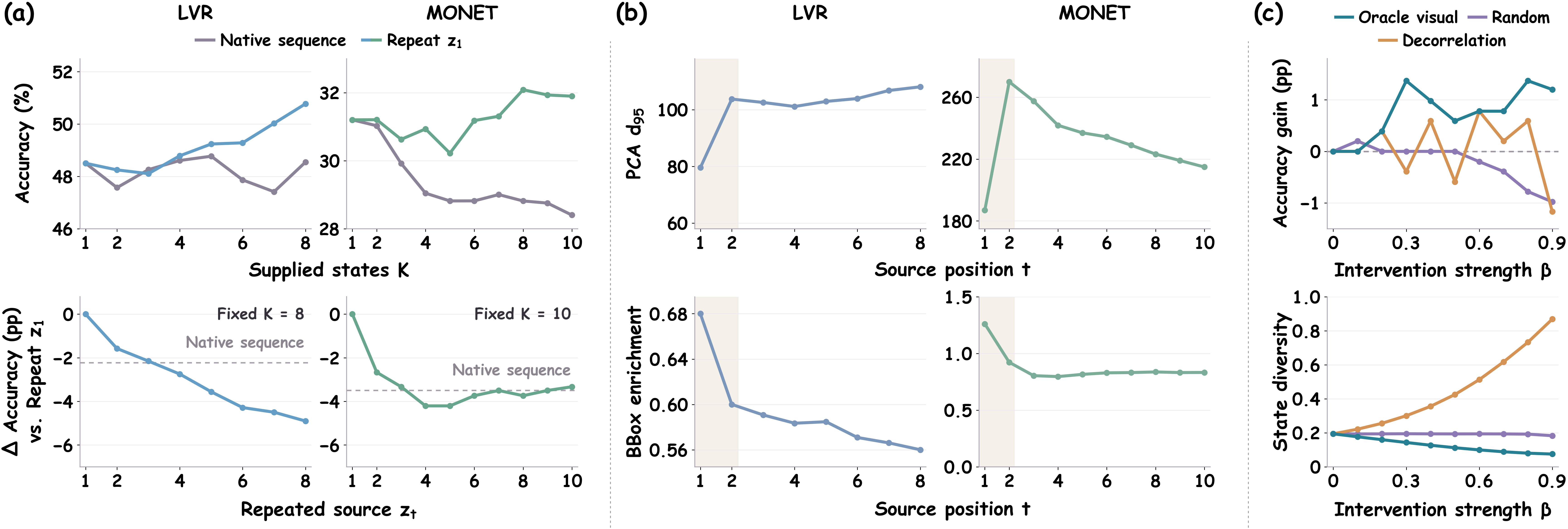}
    \vspace{-20pt}
    \caption{\textbf{State evolution, reuse, and steering in LVR-3B and MONET-7B.}
    (a) Native sequence versus repeated $z_1$ across supplied
    lengths (top), and accuracy change from repeating later
    states instead of $z_1$ at fixed budgets (bottom).
    (b) Across-sample dimensionality
    (PCA $d_{95}$, top) and BBox enrichment (bottom)
    across latent steps.
    (c) LVR-3B accuracy gains (top) and
    state diversity (bottom) under oracle BBox-guided
    visual steering, random steering, and decorrelation.}
    \vspace{-10pt}
    \label{fig:evolution}
\end{figure*}

\textcolor{black}{The upper panels of Fig.~\ref{fig:evolution}a show that
repeating $z_1$ outperforms the native sequence across most
lengths and at both full budgets, demonstrating that native
cross-step evolution brings no benefit over reusing a fixed
state. However, the lower panels show that the choice of
repeated state matters, as every later state $z_t$ ($t>1$)
yields lower accuracy than $z_1$ at the same fixed budget (additional controls in
Appendix~\ref{app:fixed_feedback_controls}). Even reusing $z_2$
already lowers accuracy by about 1.6 and 2.7 points in LVR and
MONET, respectively, motivating an analysis of how cross-step
updates alter the initial state.}

\paragraph{Subsequent updates expand dimensionality while weakening visual focus.}
To understand why accuracy drops as soon as $z_2$ is reused, we
test whether later states suffer from representation collapse or
instead expand in dimensionality while losing the visual
alignment identified in Sec.~\ref{sec:formation}. Across the
cached trajectories on the same examples, we measure
across-sample dimensionality via PCA $d_{95}$ (the number of
principal components explaining at least 95\% of variance at
step $t$) and spatial focus on the target region via final-layer
BBox enrichment (Sec.~\ref{sec:formation}).
Metric definitions and preprocessing are detailed in
Appendix~\ref{app:v2_diagnostic_evolution}.

The upper panels of Fig.~\ref{fig:evolution}b show that later
states do not collapse; instead, both models undergo a sharp
dimensional expansion from $z_1$ to $z_2$, after which LVR
maintains this elevated profile while MONET gradually contracts
yet remains above its initial level. In the lower panels, this
expansion coincides with weaker visual focus, as final-layer
BBox enrichment declines overall from approximately 0.68 at
$z_1$ to 0.56 at $z_8$ in LVR and drops over the first few steps
from roughly 1.26 before stabilizing around 0.8 in MONET. Thus,
$z_1$ concentrates across-sample variation within a more compact
subspace while maintaining stronger target focus, whereas later
states spread across more directions and attend less to the
target region. These observations motivate testing whether
steering updates toward a subspace anchored by relevant visual
features, rather than merely increasing state diversity, can
preserve state quality during evolution.

\paragraph{Restoring visual grounding, rather than diversity, improves later states.}
To test whether evolving states benefit causally from visual
grounding or from cross-step diversity, we perform opposing
interventions on frozen LVR across four-step trajectories on
the same 512 examples.
Using oracle target-box annotations that are unavailable at
inference, we steer each generated state toward the visual
subspace of target-region image tokens before supplying it to
subsequent steps. We contrast this oracle visual steering with
a random-subspace control across intervention strengths $\beta$,
and with a decorrelation intervention that pushes each state
away from preceding states to increase cross-step cosine
dissimilarity.
Intervention details and control protocols are provided in
Appendix~\ref{app:evolution_steering}.

In Fig.~\ref{fig:evolution}c (top), oracle visual steering
improves accuracy for $\beta \ge 0.3$, by up to about 1.4 points,
even as state diversity steadily declines, whereas random-subspace
steering fails to reproduce these gains and degrades accuracy at
larger $\beta$. Strength-matched controls using random directions
or mismatched images likewise fall short of target-region
guidance, indicating that the gain is not explained by
perturbation magnitude alone. Conversely, decorrelation sharply increases diversity (bottom)
yet yields erratic, inconsistent accuracy changes.

\takeaway[2]{State variation is not required for effective evolution; keeping states visually grounded, unlike increasing diversity alone, improves accuracy.}


\subsection{State Readout}
\label{sec:readout}
State readout maps the image, question, and latent
sequence $Z_K$ to the final answer. Beyond the visual recovery
emphasized in prior work~\cite{zhang2026visual}, visual
reasoning also requires task information specifying what to
infer from visual evidence. We therefore examine whether native
readout extracts task-relevant information from cached states
when the question is withheld, and whether improved readout can
recover task cues that native decoding leaves unused.

\paragraph{Withholding the question exposes a task-conditioned readout deficit.}
To test what native readout extracts from $Z_K$ without direct
access to the original inputs, we decouple state generation from
answer prediction (Appendix~\ref{app:v2_diagnostic_readout}).
Using frozen LVR-3B and MONET-7B checkpoints on the same 512
examples, we cache ten latent states per example from full
image--question inputs. During answer prediction, each model
receives these cached states and a generic answer prompt with
either the image or the question withheld (retaining cached
interleaved text for MONET). As explicit references, we also
supply annotated BBox image embeddings when withholding the
image, or answer-free task-plan tokens specifying what to
inspect without revealing observations, deductions, or answers
when withholding the question
(Appendix~\ref{app:task_plan_generation}).

\begin{figure*}[t]
    \centering
    \includegraphics[width=\linewidth,height=5cm]{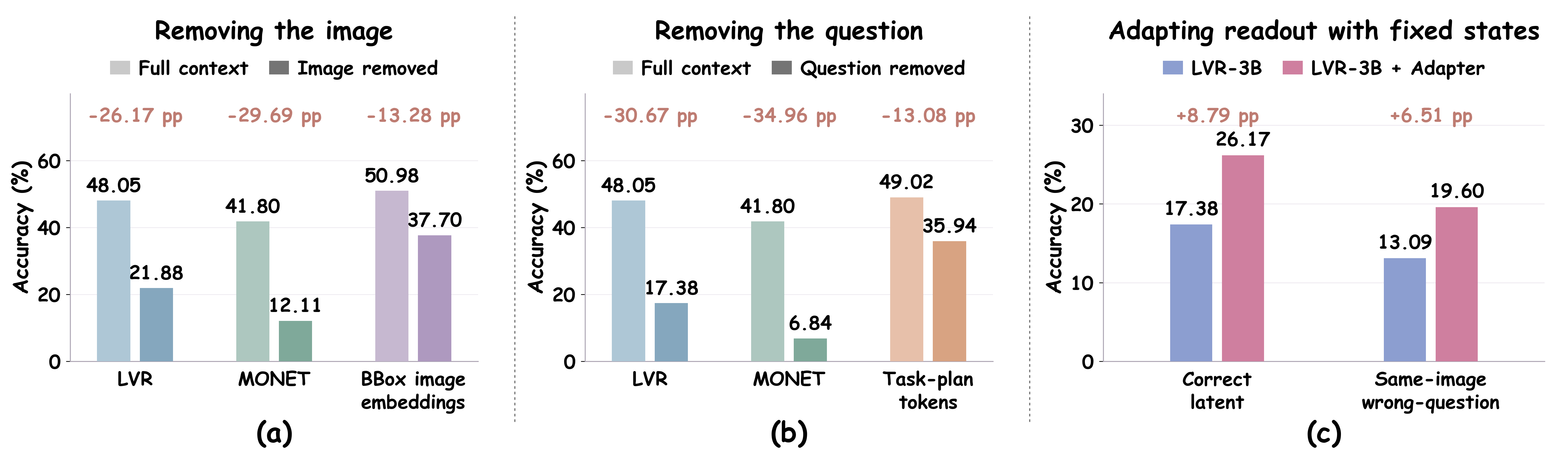}
    \vspace{-20pt}
    \caption{\textbf{Visual and task-conditioned state readout.}
    (a,b) Context removal with latent, task-plan, or visual inputs.
    (c) LVR-3B versus LVR-3B + Adapter with the backbone
    and source caches fixed, evaluated without the question
    using correct, same-image wrong-question, or no latent states.
    Adapter results average three training seeds.}
    \vspace{-10pt}
    \label{fig:readout}
\end{figure*}

Withholding either input sharply degrades accuracy from the
full-input baselines of 48.05\% in LVR and 41.80\% in MONET.
Without the image (Fig.~\ref{fig:readout}a), LVR and MONET
reach only 21.88\% and 12.11\%, whereas supplying BBox image
embeddings instead retains 37.70\% on average across both
models; without the question (Fig.~\ref{fig:readout}b),
accuracy falls further to 17.38\% and 6.84\%, whereas
task-plan tokens retain 35.94\%.
Thus, even when the image
remains visible and $Z_K$ was generated from the full
image--question pair, native readout struggles to answer the
question.

\paragraph{Cached states retain task cues that native readout under-utilizes.}
To test whether task cues are absent from $Z_K$ or encoded yet
under-utilized, we freeze both LVR-3B and its cached states
and train only a lightweight adapter (\emph{LVR-3B + Adapter})
to transform those states before answer prediction
(Appendix~\ref{app:trained_state_readout}). Using four-state
caches generated from full image--question inputs, we optimize
the adapter for eight epochs with answer supervision on 512
disjoint GQA examples under the question-withheld setting,
supplying only the image, adapted states, and a generic answer
prompt. We then evaluate on the same 512 diagnostic examples
across three training seeds, comparing native and adapted
readout on identical caches across correct states, same-image
mismatched states from another question, and a no-latent
baseline.

In Fig.~\ref{fig:readout}c, adapting readout on the same
correct caches raises question-withheld accuracy from 17.38\%
to 26.17\%, whereas an adapter trained on shuffled caches
reaches only 18.03\% (Table~\ref{tab:adapter_readout}) and the
no-latent baseline stays at 0.59\%. The margin of correct over
same-image mismatched states also widens from 4.30 to 6.58
percentage points, indicating that adapted readout extracts
additional question-specific cues rather than relying on image
or dataset priors alone. Because both the backbone and cached
states remain fixed, these gains show that $Z_K$ encodes more
task-relevant information than native decoding uses, making
task readout an explicit bottleneck alongside visual recovery.

\takeaway[3]{Beyond visual recovery, latent states encode task-relevant cues that native decoding fails to fully extract, making task-conditioned readout an explicit bottleneck.}

\vspace{-5pt}
\vspace{-3pt}

\section{Method}
\label{sec:method}
\label{sec:method_overview}

\begin{figure*}[t]
    \centering
    \includegraphics[width=\linewidth]{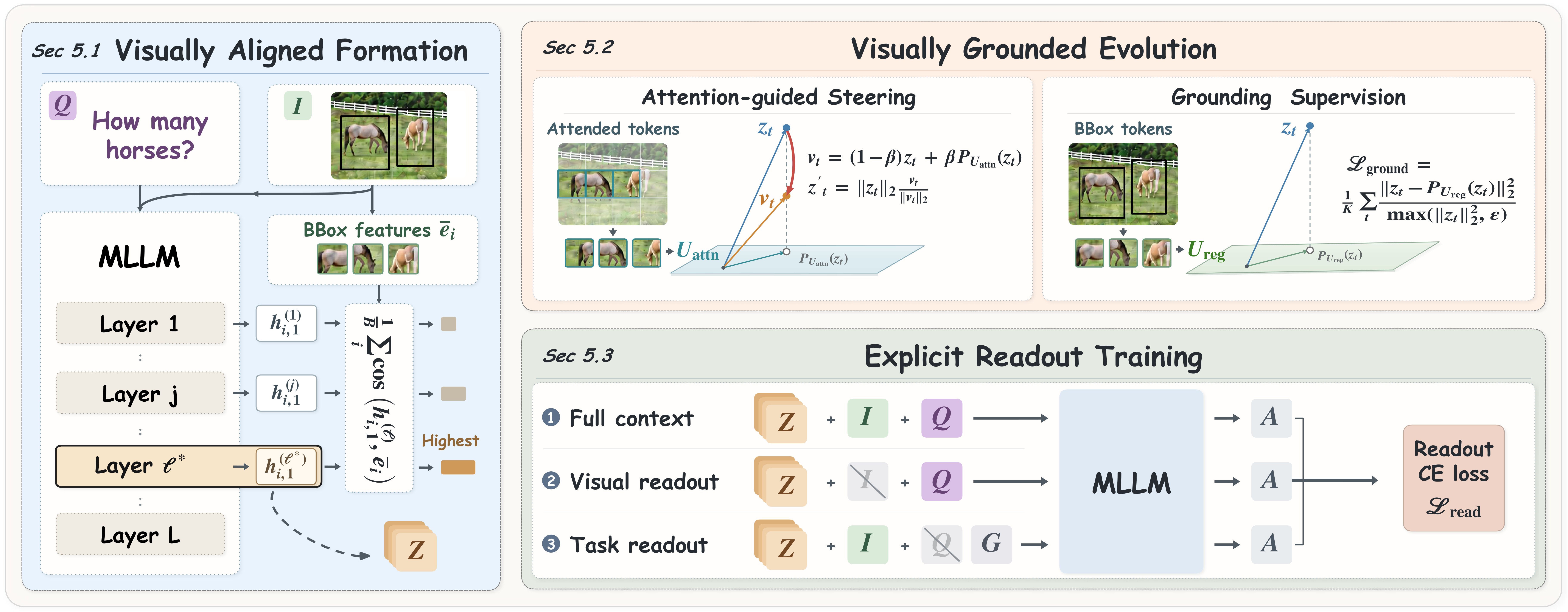}
    \vspace{-20pt}
    \caption{\textbf{Overview of LIFE-VLR.}
    For formation, latent states come
    from the non-final layer $\ell^\star$ best aligned with BBox features.
    For evolution, attention-guided
    steering ($U_{\mathrm{attn}}$) and grounding supervision
    ($\mathcal{L}_{\mathrm{ground}}$) preserve visual focus.
    For readout, a staged curriculum
    withholds $I$ or replaces $Q$ with a generic instruction $G$ so the
    MLLM learns to extract both visual and task cues from $Z$.}
    \vspace{-10pt}
    \label{fig:pipline}
\end{figure*}

Our preceding analyses show that visual and task signals break
down at different stages of latent reasoning. Visual alignment
drops at the final layer during state formation and weakens
further across state evolution, whereas task cues remain
encoded in the latent states yet are under-utilized during
native readout. Guided by these findings, we develop
\textbf{LIFE-VLR} (Fig.~\ref{fig:pipline}), a unified training
procedure that selects visually aligned source layers, anchors
evolving states to visual subspaces, and strengthens both
visual and task readout via a staged curriculum, requiring
neither region annotations nor extra parameters at inference.
We use LVR for exposition; MONET applies the same interventions while retaining its distillation objectives (Appendix~\ref{app:v2_monet}).

\subsection{Visually Aligned State Formation}
\label{sec:method_generation}
\label{sec:method_formation}

Rather than defaulting to the final decoder layer $L$, we
form states from an earlier, visually aligned layer. For
training example $i$, let $\bar e_i$ denote the mean of its
bounding-box-selected image embeddings from the frozen visual
encoder and merger. The initial forward pass on the full
image--question prompt produces $h_{i,1}^{(\ell)}$, the raw
output of decoder layer $\ell$ at the first latent-prediction
position. For each local microbatch of $B$ examples, we select
the non-final layer with the highest mean cosine similarity:
\begingroup
\small
\begin{equation}
\ell^\star =
\underset{1\leq\ell<L}{\arg\max}\;
\frac{1}{B}\sum_{i=1}^{B}
\cos\!\left(h_{i,1}^{(\ell)},\bar e_i\right).
\label{eq:v2_formation}
\end{equation}
\endgroup
We select $\ell^\star$ anew per microbatch and hold it fixed
across the resulting latent trajectory. All $L$ decoder layers
still execute, so selection changes the state source rather
than decoder depth. At inference, we fix $\ell^\star$ to the
layer most frequently selected during training, determining
each backbone's source in a single run without separate
layer-wise search (Appendix~\ref{app:v2_inference}).
Alternative selection criteria are evaluated in
Appendix~\ref{app:source_selection_ablation}.

\subsection{Visually Grounded State Evolution}
\label{sec:method_evolution}

\begingroup

While steering states toward target visual regions
counteracts drift across updates
(Sec.~\ref{sec:state_evolution}), ground-truth regions are
unavailable at inference. We therefore pair
\emph{attention-guided steering}, which pulls evolving states
toward model-attended image tokens, with a training-time
\emph{region-grounding loss} that aligns raw predictions with
ground-truth regions. Specifically, upon reading $(I, Q)$ at
the first latent-prediction position, we use the MLLM's
attention to select relevant original-image tokens and
construct an orthonormal basis $U_{\mathrm{attn}}$ from their
embeddings, defining a fixed visual subspace for the
trajectory. Writing $P_U(z)=U(U^\top z)$ for orthogonal
projection onto the subspace spanned by $U$, we compute
the steered state $z'_t$ at step $t$ as
\begingroup
\small
\begin{equation}
\begin{aligned}
z_t &= F_\theta^{(\ell^\star)}(I,Q,Z_{t-1}),\\
v_t &= (1-\beta)z_t
       +\beta P_{U_{\mathrm{attn}}}(z_t),\\
z'_t &= \|z_t\|_2\,\frac{v_t}{\|v_t\|_2},
\end{aligned}
\label{eq:v2_feedback}
\end{equation}
\endgroup
where $F_\theta^{(\ell^\star)}$ returns the raw prediction
$z_t$ from layer $\ell^\star$ (starting at the latent-start
marker for $t=1$), and the steered state $z'_t$ is appended
to $Z$ in place of $z_t$.
The coefficient $\beta$ controls steering strength,
with the mixture $v_t$ preserving the component of
$z_t$ inside the visual subspace and scaling its off-subspace
component by $1-\beta$. Rescaling restores the original norm
$\|z_t\|_2$, so steering adjusts direction without altering
state magnitude.

Because steering only adjusts states after $z_t$ is produced,
we also supervise $z_t$ directly. From bounding-box-selected
image embeddings, we construct a target-region basis
$U_{\mathrm{reg}}$ and penalize the normalized residual energy
of $z_t$ outside this subspace:
\begingroup
\small
\setlength{\abovedisplayskip}{5pt}
\setlength{\belowdisplayskip}{5pt}
\setlength{\abovedisplayshortskip}{3pt}
\setlength{\belowdisplayshortskip}{5pt}
\begin{equation}
\mathcal{L}_{\mathrm{ground}} =
\frac{1}{K}\sum_{t=1}^{K}
\frac{
  \|z_t-P_{U_{\mathrm{reg}}}(z_t)\|_2^2
}{
  \max(\|z_t\|_2^2,\epsilon)
}.
\label{eq:v2_grounding}
\end{equation}
\endgroup
Here, $\epsilon > 0$ ensures numerical stability. The region
basis $U_{\mathrm{reg}}$ provides training-time supervision,
whereas the attention basis $U_{\mathrm{attn}}$ steers state
updates during both training and inference (details in
Appendix~\ref{app:v2_basis_math}).

\endgroup

\subsection{Explicit State Readout Training}
\label{sec:method_readout}

When $(I, Q)$ remain fully visible during answer decoding,
the reader can attend directly to the raw inputs rather than
relying on $Z$ (Sec.~\ref{sec:readout}). To train the reader
to extract both visual and task cues from $Z$, we always
generate $Z$ from the full $(I, Q)$ during training, but
condition answer decoding on $(C, Z)$, where the external
context $C$ available to the reader progresses through three
stages:
\begingroup
\setlength{\abovedisplayskip}{5pt}
\setlength{\belowdisplayskip}{5pt}
\setlength{\abovedisplayshortskip}{3pt}
\setlength{\belowdisplayshortskip}{5pt}
\begin{equation}
C:\quad
(I,Q)\ \longrightarrow\ Q\ \longrightarrow\ (I,G).
\label{eq:v2_curriculum}
\end{equation}
\endgroup
Here, $G$ is a fixed, question-free instruction prompting the
reader to answer using $Z$. Training first uses $(I, Q)$ to
establish answer prediction with $Z$, then removes the image to
encourage visual readout from $Z$, and finally restores the
image while replacing $Q$ with $G$ to encourage task readout.
Input withholding is applied only during training; at
inference, the reader receives the full $(I, Q)$ alongside
$Z$.

\subsection{Training Objective}
\label{sec:method_objective}
We combine grounding supervision under the full context
$(I, Q)$ with answer supervision under the scheduled reader
context $C$. For answer tokens $A=(a_1,\ldots,a_{|A|})$, with
$a_{<j}$ denoting preceding answer tokens, the training
objective is
\begingroup
\small
\setlength{\abovedisplayskip}{5pt}
\setlength{\belowdisplayskip}{5pt}
\setlength{\abovedisplayshortskip}{3pt}
\setlength{\belowdisplayshortskip}{5pt}
\begin{equation}
\begin{aligned}
\mathcal{L}_{\mathrm{read}}
&=
-\frac{1}{|A|}\sum_{j=1}^{|A|}
\log p_\theta
\left(a_j\mid C,\operatorname{sg}(Z),a_{<j}\right),\\
\mathcal{L}
&=
\mathcal{L}_{\mathrm{read}}
+\lambda\mathcal{L}_{\mathrm{ground}},
\end{aligned}
\label{eq:v2_objective}
\end{equation}
\endgroup
with both losses averaged over examples and $\lambda$
weighting grounding supervision. The stop-gradient operator
$\operatorname{sg}$ treats $Z$ as fixed input, so
$\mathcal{L}_{\mathrm{read}}$ updates the shared parameters
$\theta$ through the readout pass without backpropagating
through state generation.

To compute $\mathcal{L}$, we first generate $Z$ without
retaining the generation graph. Using the saved states, we
then recompute $z_t$ under $(I, Q)$ with gradients for
$\mathcal{L}_{\mathrm{ground}}$ and evaluate
$\mathcal{L}_{\mathrm{read}}$ under $C$, treating $\ell^\star$
and the visual bases as constants. For each readout pass, the
input sequence and KV cache are rebuilt under $C$ to exclude
any withheld image or question. Further details are provided
in Appendices~\ref{app:v2_training} and~\ref{app:v2_combo}.

\section{Experiments}
\label{sec:experiments}
\subsection{Experimental Setup}
\label{sec:setup}
\label{sec:experimental_setup}

We evaluate LIFE-VLR within two visual latent reasoning
frameworks, LVR~\cite{li2025latent} and
MONET~\cite{wang2025monet}, across three MLLM backbones
(Qwen2.5-VL-3B, Qwen3-VL-8B, and Qwen3.5-27B). Following
each official implementation, we train LVR on
Visual-CoT~\cite{shao2024visualcot} and MONET on
Monet-SFT-125K, comparing MLLMs trained with LIFE-VLR
against those trained with the official recipe under
identical data and optimization budgets. Across all
frameworks and backbones, we set steering strength
$\beta=0.2$, grounding-loss weight $\lambda=0.1$, and
numerical stabilizer $\epsilon=10^{-8}$. Implementation
details are provided in Appendix~\ref{app:v2_implementation}.

We report accuracy on nine benchmarks spanning high-resolution
understanding, fine-grained perception, and visual reasoning,
comprising \mbox{VStarBench}~\cite{wu2024vstar}, HRBench4K,
HRBench8K~\cite{hrbench}, BLINK~\cite{fu2024blink},
MMVP~\cite{tong2024eyes}, MME-RealWorld~\cite{zhang2024mme},
MMStar~\cite{chen2024we},
VisualPuzzles~\cite{song2025visualpuzzles}, and
CVBench~\cite{tong2024cambrian1}. Evaluation uses
VLMEvalKit~\cite{duan2024vlmevalkit} with its
benchmark-specific prompting, answer extraction, and scoring
procedures. At inference, LVR uses $K=4$ latent steps and
MONET uses eight latent positions per segment. Further
details are provided in Appendix~\ref{app:v2_evaluation}.


\begin{table*}[t]
\centering
\vspace{-5pt}
\caption{\textbf{Results across nine visual reasoning benchmarks.}
Scores are accuracies (\%).
Bold indicates the better result within each baseline/+Ours pair;
improvements are measured in percentage points.}
\label{tab:main_results}

\begingroup
\definecolor{MainTableGain}{RGB}{0,145,70}
\definecolor{MainTableMuted}{RGB}{110,110,110}

\providecommand{\MainGain}[1]{}
\renewcommand{\MainGain}[1]{%
    \textcolor{MainTableGain}{\textit{+#1}}}

\fontsize{8.5}{9.2}\selectfont
\setlength{\tabcolsep}{1.5pt}
\setlength{\aboverulesep}{1pt}
\setlength{\belowrulesep}{1pt}
\renewcommand{\arraystretch}{0.94}

\begin{tabular*}{\textwidth}{
    @{\extracolsep{\fill}}cl*{9}{c}@{}
}
\toprule
\textbf{Backbone}
& \textbf{Method}
& \textbf{V$^{\star}$}
& \shortstack{\textbf{HRBench}\\\textbf{4K}}
& \shortstack{\textbf{HRBench}\\\textbf{8K}}
& \textbf{BLINK}
& \textbf{MMVP}
& \shortstack{\textbf{MME-}\\\textbf{RealWorld}}
& \textbf{MMStar}
& \shortstack{\textbf{Visual}\\\textbf{Puzzles}}
& \textbf{CVBench} \\
\midrule

\multirow{6}{*}{\shortstack[c]{Qwen2.5-VL\\3B}}
& LVR
& 66.49 & 56.38 & 52.75 & 43.66 & 61.33
& 47.65 & 49.67 & 28.42 & 48.22 \\
& LVR + \textbf{Ours}
& \textbf{73.30} & \textbf{63.25} & \textbf{58.75}
& \textbf{47.03} & \textbf{64.00} & \textbf{49.19}
& \textbf{53.67} & \textbf{31.59} & \textbf{60.44} \\
& \textcolor{MainTableMuted}{\textit{Improvement}}
& \MainGain{6.81} & \MainGain{6.87} & \MainGain{6.00}
& \MainGain{3.37} & \MainGain{2.67} & \MainGain{1.54}
& \MainGain{4.00} & \MainGain{3.17} & \MainGain{12.22} \\
\addlinespace[1.5pt]
& MONET
& 64.92 & 62.75 & 58.13 & 39.24 & 56.33
& 51.73 & 45.60 & 22.26 & 49.08 \\
& MONET + \textbf{Ours}
& \textbf{69.11} & \textbf{64.62} & \textbf{60.63}
& \textbf{42.40} & \textbf{58.67} & \textbf{54.20}
& \textbf{47.40} & \textbf{24.66} & \textbf{56.20} \\
& \textcolor{MainTableMuted}{\textit{Improvement}}
& \MainGain{4.19} & \MainGain{1.87} & \MainGain{2.50}
& \MainGain{3.16} & \MainGain{2.34} & \MainGain{2.47}
& \MainGain{1.80} & \MainGain{2.40} & \MainGain{7.12} \\
\midrule

\multirow{6}{*}{\shortstack[c]{Qwen3-VL\\8B}}
& LVR
& 76.96 & 68.00 & 63.25 & 50.87 & 71.00
& 48.66 & 60.40 & 26.63 & 70.60 \\
& LVR + \textbf{Ours}
& \textbf{80.63} & \textbf{74.38} & \textbf{69.12}
& \textbf{58.97} & \textbf{77.00} & \textbf{60.21}
& \textbf{61.07} & \textbf{28.25} & \textbf{71.45} \\
& \textcolor{MainTableMuted}{\textit{Improvement}}
& \MainGain{3.67} & \MainGain{6.38} & \MainGain{5.87}
& \MainGain{8.10} & \MainGain{6.00} & \MainGain{11.55}
& \MainGain{0.67} & \MainGain{1.62} & \MainGain{0.85} \\
\addlinespace[1.5pt]
& MONET
& 74.87 & 63.88 & 60.88 & 52.18 & 65.00
& \textbf{49.75} & 60.67 & 26.88 & 71.08 \\
& MONET + \textbf{Ours}
& \textbf{77.49} & \textbf{70.50} & \textbf{67.50}
& \textbf{55.71} & \textbf{68.67} & 48.97
& \textbf{61.80} & \textbf{27.74} & \textbf{77.75} \\
& \textcolor{MainTableMuted}{\textit{Improvement}}
& \MainGain{2.62} & \MainGain{6.62} & \MainGain{6.62}
& \MainGain{3.53} & \MainGain{3.67}
& \textit{\ensuremath{-}0.78}
& \MainGain{1.13} & \MainGain{0.86} & \MainGain{6.67} \\
\midrule

\multirow{6}{*}{\shortstack[c]{Qwen3.5\\27B}}
& LVR
& 79.06 & 72.75 & 66.12 & 60.92 & 77.67
& 66.56 & 63.87 & 28.94 & 66.59 \\
& LVR + \textbf{Ours}
& \textbf{82.72} & \textbf{75.75} & \textbf{70.50}
& \textbf{63.97} & \textbf{80.33} & \textbf{70.27}
& \textbf{65.13} & \textbf{34.50} & \textbf{83.94} \\
& \textcolor{MainTableMuted}{\textit{Improvement}}
& \MainGain{3.66} & \MainGain{3.00} & \MainGain{4.38}
& \MainGain{3.05} & \MainGain{2.66} & \MainGain{3.71}
& \MainGain{1.26} & \MainGain{5.56} & \MainGain{17.35} \\
\addlinespace[1.5pt]
& MONET
& 85.86 & 80.38 & 74.62 & \textbf{64.70} & 72.00
& 70.39 & \textbf{68.87} & 40.67 & 72.93 \\
& MONET + \textbf{Ours}
& \textbf{87.43} & \textbf{82.12} & \textbf{77.75}
& 64.28 & \textbf{76.00} & \textbf{71.79} & 67.93
& \textbf{42.38} & \textbf{75.66} \\
& \textcolor{MainTableMuted}{\textit{Improvement}}
& \MainGain{1.57} & \MainGain{1.74} & \MainGain{3.13}
& \textit{\ensuremath{-}0.42}
& \MainGain{4.00} & \MainGain{1.40}
& \textit{\ensuremath{-}0.94}
& \MainGain{1.71} & \MainGain{2.73} \\
\bottomrule
\end{tabular*}
\endgroup
\end{table*}

\begin{table}[t]
\centering
\caption{\textbf{SFT gains persist after RL.}
Each pair uses the same framework-specific RL procedure;
bold marks the higher score.
}
\label{tab:rl_validation}
\vspace{-2pt}

\begingroup
\fontsize{8.5}{9.5}\selectfont
\setlength{\tabcolsep}{2.5pt}
\setlength{\aboverulesep}{1.5pt}
\setlength{\belowrulesep}{1.5pt}
\renewcommand{\arraystretch}{1.1}

\begin{tabular*}{\linewidth}{
    @{\extracolsep{\fill}}l*{5}{r}@{}
}
\toprule
\textbf{Training pipeline}
& \textbf{V$^{\star}$}
& \textbf{HR4K}
& \textbf{HR8K}
& \textbf{BLINK}
& \textbf{MMVP} \\
\midrule

LVR $\rightarrow$ RL
& 68.59 & 60.12 & 54.62 & 46.87 & 67.67 \\
(LVR + \textbf{Ours}) $\rightarrow$ RL
& \textbf{73.30} & \textbf{65.63} & \textbf{58.88}
& \textbf{49.34} & \textbf{68.33} \\
\addlinespace[1.5pt]

MONET $\rightarrow$ RL
& 68.59 & 62.88 & 57.25 & 41.35 & 58.33 \\
(MONET + \textbf{Ours}) $\rightarrow$ RL
& \textbf{71.73} & \textbf{65.12} & \textbf{60.25}
& \textbf{43.66} & \textbf{60.67} \\
\bottomrule
\end{tabular*}
\endgroup
\vspace{-2pt}
\end{table}

\begin{table}[t]
\centering
\caption{\textbf{Component ablations.}
Accuracy (\%) on Qwen2.5-VL-3B.
F: source selection; E: visual preservation; R: readout training.
Runs follow official LVR settings
(Appendix~\ref{app:component_only}).
Bold/underline indicate best/second-best scores.}
\vspace{-2pt}
\label{tab:component_validation}

\begingroup
\fontsize{8.5}{9.5}\selectfont
\setlength{\tabcolsep}{2pt}
\setlength{\aboverulesep}{1.5pt}
\setlength{\belowrulesep}{1.5pt}
\renewcommand{\arraystretch}{1.02}

\begin{tabular*}{\linewidth}{
    @{}
    *{3}{>{\centering\arraybackslash}p{0.45cm}}
    @{\hspace{6pt}\extracolsep{\fill}}
    *{5}{c}
    @{}
}
\toprule
\multicolumn{3}{c}{\textbf{Components}}
& \multirow{2}{*}{\textbf{V$^{\star}$}}
& \multirow{2}{*}{\textbf{HR4K}}
& \multirow{2}{*}{\textbf{HR8K}}
& \multirow{2}{*}{\textbf{BLINK}}
& \multirow{2}{*}{\textbf{MMVP}} \\
\cmidrule(lr){1-3}
\textbf{F} & \textbf{E} & \textbf{R}
& & & & & \\
\midrule

 &  & 
& 66.49 & 56.38 & 52.75 & 43.66 & 61.33 \\
\midrule

$\checkmark$ &  & 
& 71.20 & 62.50 & 57.00 & \underline{47.03} & 64.00 \\
 &  & $\checkmark$
& 70.16 & 59.88 & 55.88 & 45.87 & 62.33 \\
 & $\checkmark$ & 
& 69.63 & 61.13 & 57.00 & 45.92 & 63.33 \\

$\checkmark$ &  & $\checkmark$
& 71.20 & 63.00 & \underline{58.25}
& \underline{47.03} & 64.00 \\
$\checkmark$ & $\checkmark$ & 
& \textbf{74.35} & 62.50 & \underline{58.25}
& \textbf{47.92} & \underline{65.00} \\
 & $\checkmark$ & $\checkmark$
& 71.20 & \underline{63.13} & 57.88
& 46.55 & \textbf{65.67} \\

$\checkmark$ & $\checkmark$ & $\checkmark$
& \underline{73.30} & \textbf{63.25} & \textbf{58.75}
& \underline{47.03} & 64.00 \\
\bottomrule
\end{tabular*}
\endgroup
\vspace{-5pt}
\end{table}

\vspace{-5pt}

\subsection{Results}
\label{sec:v2_main_results}
\textbf{Main results.}
Across three backbones and nine benchmarks under matched
budgets, LIFE-VLR (\textbf{Ours} in tables) outperforms the
official LVR and MONET recipes in 51 of 54 comparisons, with
the three exceptions each within 1 percentage point (pp)
(Table~\ref{tab:main_results}). LVR improves on all nine
benchmarks at every scale, while MONET improves on nine,
eight, and seven benchmarks at 3B, 8B, and 27B. Gains span
high-resolution and fine-grained perception (winning all six
framework--backbone pairs on VStarBench, HRBench4K,
HRBench8K, and MMVP, and reaching $+11.55$~pp on
MME-RealWorld) as well as visual reasoning, reaching
$+12.22$~pp and $+17.35$~pp on CVBench with 3B and 27B LVR.

\noindent\textbf{Persistence after RL.}
To test whether these SFT gains persist under subsequent
reinforcement learning (RL), we apply continued RL to
Qwen2.5-VL-3B under limited, framework-specific budgets
matched within each baseline/LIFE-VLR pair
(Table~\ref{tab:rl_validation}). RL further improves our
checkpoints on four of five benchmarks in each framework,
while LIFE-VLR remains ahead of the corresponding RL
baselines on all five (by up to $+5.51$~pp in LVR and
$+3.14$~pp in MONET).

\noindent\textbf{Component ablations.}
Evaluating state formation (F), state evolution (E), and
readout training (R) on LVR with Qwen2.5-VL-3B shows that
each component alone improves over the baseline on all five
benchmarks (Table~\ref{tab:component_validation}). Combining
all three matches or exceeds every single-component variant
and achieves the highest accuracy on HRBench4K ($63.25\%$)
and HRBench8K ($58.75\%$), whereas pairwise combinations
(F+E or E+R) lead on VStarBench, BLINK, and MMVP,
indicating that the components interact rather than combine
additively.

\section{Conclusion}
\label{sec:conclusion}
We revisit bottlenecks in visual latent reasoning across
state formation, evolution, and readout. Our analysis shows
that the final decoder layer is not consistently the best
state source, preserving visual evidence during evolution may
matter more than increasing state diversity, and task
information retained in latent states can remain underused by
the receiving MLLM. Building on these findings, LIFE-VLR
combines visually aligned source selection, grounded state
evolution, and staged readout training, requiring neither
region annotations nor extra parameters at inference. Across
LVR and MONET, LIFE-VLR improves accuracy in 51 of 54
comparisons spanning 3B--27B backbones and nine benchmarks,
and retains its advantage under continued reinforcement
learning.

\clearpage
{\small
\bibliographystyle{ieeenat_fullname}
\bibliography{references}}

\clearpage
\appendix

\section{Overview of the Appendix}
\label{app:v2_overview}

The appendix connects the diagnostic evidence to the training
procedure and its evaluation. We first give the settings,
protocols, and controls behind the formation, evolution, and
readout analyses, then describe how the interventions are
implemented in LVR and MONET. Training and post-training analyses revisit the model's behavior.
After specifying the evaluation protocol, we assess the resulting
procedure through ablations, repeated runs, and cost measurements.
Prompt templates and qualitative examples provide the context
needed to interpret these comparisons.

\begingroup
\small
\setlength{\parindent}{0pt}

\newcommand{\AppendixEntry}[2]{%
    \par\addvspace{5pt}\noindent
    \hyperref[#1]{\textbf{\ref*{#1}\quad #2}}%
    \nobreak\dotfill\pageref{#1}\par
}

\newcommand{\AppendixSubentry}[2]{%
    \par\addvspace{2pt}\noindent
    \hspace*{10pt}%
    \hyperref[#1]{\ref*{#1}\quad #2}%
    \nobreak\dotfill\pageref{#1}\par
}

\noindent\textbf{Contents}\hfill\textbf{Page}\par

\AppendixEntry{app:v2_scope}{Limitations and Broader Impact}

\AppendixEntry{app:v2_analysis_settings}{Experimental Details of the Analyses}
\AppendixSubentry{app:v2_models}{Models}
\AppendixSubentry{app:v2_data}{Data}
\AppendixSubentry{app:v2_diagnostic_scoring}{Scoring and Comparison Settings}

\AppendixEntry{app:v2_diagnostics}{Diagnostic Protocol Details}
\AppendixSubentry{app:v2_diagnostic_formation}{State Formation}
\AppendixSubentry{app:v2_diagnostic_evolution}{State Evolution}
\AppendixSubentry{app:v2_diagnostic_readout}{State Readout}

\AppendixEntry{app:additional_diagnostic_controls}{Additional Diagnostic Controls}
\AppendixSubentry{app:formation_crossed_ablation}{Visual Supervision and State Source}
\AppendixSubentry{app:formation_normalization}{Final-Layer Normalization}
\AppendixSubentry{app:fixed_feedback_controls}{Controls for Fixed-State Repetition}
\AppendixSubentry{app:steering_matched_controls}{Strength-Matched Controls for Visual Steering}
\AppendixSubentry{app:adapter_readout_controls}{Controls for Adapter-Based Readout}

\AppendixEntry{app:method_implementation}{Method Implementation Details}
\AppendixSubentry{app:v2_basis_math}{Source Selection and Visual Bases}
\AppendixSubentry{app:v2_combo}{LVR Training Procedure}
\AppendixSubentry{app:v2_monet}{MONET Training Procedure}
\AppendixSubentry{app:v2_inference}{Inference and Backbone Adaptation}
\AppendixSubentry{app:rl_implementation}{Reinforcement-Learning Implementation}

\AppendixEntry{app:v2_implementation}{Training Details}
\AppendixSubentry{app:v2_training}{LVR Supervised Training}
\AppendixSubentry{app:monet_training_configuration}{MONET Supervised Training}
\AppendixSubentry{app:v2_rl_training}{Reinforcement Learning}

\AppendixEntry{app:v2_training_dynamics}{Training Dynamics}

\AppendixEntry{app:trained_model_diagnostics}{Analysis of the Model after Training}
\AppendixSubentry{app:post_training_grounding}{Visual Grounding and State Diversity}
\AppendixSubentry{app:post_training_readout}{Task Readout}

\AppendixEntry{app:v2_evaluation}{Evaluation Details}
\AppendixSubentry{app:eval_benchmarks}{Benchmarks and Evaluation Toolkit}
\AppendixSubentry{app:eval_decoding}{Models, Inputs, and Decoding}
\AppendixSubentry{app:eval_scoring}{Scoring and Reporting}

\AppendixEntry{app:v2_supplementary}{Additional Ablation Study}
\AppendixSubentry{app:source_selection_ablation}{Source-Layer Selection}
\AppendixSubentry{app:v2_grounding_weight}{Grounding Weight}
\AppendixSubentry{app:feedback_strength}{Steering Strength}
\AppendixSubentry{app:readout_curriculum_ablation}{Readout Curriculum}
\AppendixSubentry{app:v2_latent_budget}{Latent Budget}

\AppendixEntry{app:v2_seed_stability}{Multi-Seed Stability}

\AppendixEntry{app:v2_efficiency}{Training Cost}

\AppendixEntry{app:v2_prompts}{Prompt Templates and Input Construction}
\AppendixSubentry{app:v2_lvr_prompt}{LVR}
\AppendixSubentry{app:v2_monet_prompt}{MONET}
\AppendixSubentry{app:restricted_prompts}{Restricted-Context Templates}
\AppendixSubentry{app:task_plan_generation}{Task-Plan Reference Generation}

\AppendixEntry{app:qualitative_examples}{Qualitative Examples and Visualizations}
\AppendixSubentry{app:qualitative_predictions}{Representative Successes}
\AppendixSubentry{app:qualitative_failures}{Remaining Challenges}

\endgroup
\section{Limitations and Broader Impact}
\label{app:v2_scope}

\subsection{Limitations}
Our study has three main limitations.
First, the controlled analyses focus on two frameworks
and a limited set of diagnostic examples.
Extending these analyses to larger models, broader data
distributions, and additional architectures would help
establish the generality of the findings.
Second, downstream improvements are less consistent at
larger model scales, and our RL validation covers one
backbone and five benchmarks under framework-specific
training recipes.
Further evaluation is needed to understand how the
identified bottlenecks and interventions behave under
broader RL settings and at larger scales.
Finally, formation, evolution, and readout provide a
systematic basis for investigating visual latent states,
but do not capture every factor affecting the effectiveness
of visual latent reasoning. Our approach does not resolve
all bottlenecks, and further limitations remain to be
identified and addressed.

\subsection{Broader Impact}
Our study provides an empirical foundation for future visual
latent reasoning research by connecting the properties of
intermediate states to their role in answer prediction.
The findings help explain what makes these states useful:
how they are formed, how they support computation across
steps, and how their information is used by the model.
By translating these findings into targeted training
interventions, we also show how the analysis can guide
improvements to existing systems.
These insights may inform the design of intermediate
representations for visual mathematical and geometric
reasoning~\cite{lu2024mathvista}, as well as spatial reasoning
about object positions, distances, and
relationships~\cite{chen2024spatialvlm}.
They may also guide future work on coordinating visual
perception and textual reasoning~\cite{lin2026dyco},
by clarifying how intermediate states can support the
interaction between the two.
Our diagnostic protocols and interventions provide concrete
starting points for studying these connections in broader
visual reasoning settings.

\section{Experimental Details of the Analyses}
\label{app:v2_analysis_settings}

\subsection{Models}
\label{app:v2_models}
The diagnostic analyses use two representative visual
latent reasoning frameworks, LVR~\cite{li2025latent} and
MONET~\cite{wang2025monet}. We train Qwen2.5-VL-3B
following the official LVR procedure and denote the
resulting checkpoint by \textbf{LVR-3B}. For MONET,
we use the officially released Qwen2.5-VL-7B checkpoint,
denoted by \textbf{MONET-7B}.

\subsection{Data}
\label{app:v2_data}
We sample 512 image--question pairs from the GQA
validation split of Visual-CoT~\cite{shao2024visualcot},
which provides answers, reasoning traces, and target-region
annotations. Both models use the same 512 examples, held fixed
across conditions within each comparison.

\subsection{Scoring and Comparison Settings}
\label{app:v2_diagnostic_scoring}
We score answers by normalized exact match after removing
answer-format tags and normalizing case, punctuation,
and whitespace in predictions and reference answers.
Within each experiment, all compared conditions use the
same examples, decoding settings, and scoring procedure.
Appendix~\ref{app:v2_diagnostics} specifies the latent
lengths, input modifications, and controls for each protocol.
\begin{table*}[t]
\centering
\caption{\textbf{Final performance with different formation sources.}
Accuracy (\%) on Qwen2.5-VL-3B.
LVR denotes the standard fixed-final baseline in
Table~\ref{tab:main_results}.
Range-based configurations randomly select a layer during
training and inference.
Avg.\ is the unweighted nine-benchmark mean.
Bold marks the highest score in each column.}
\label{tab:formation_source_depth}
\begingroup
\fontsize{8.5}{9.5}\selectfont
\setlength{\tabcolsep}{2.5pt}
\renewcommand{\arraystretch}{1.02}
\begin{tabular*}{\textwidth}{@{\extracolsep{\fill}}l*{10}{r}@{}}
\toprule
Source range & V$^{\star}$ & HR4K & HR8K & BLINK & MMVP
& MME-RW & MMStar & VisPuz & CVBench & Avg. \\
\midrule
LVR (fixed L36)
& 66.49 & 56.38 & 52.75 & 43.66 & 61.33
& 47.65 & 49.67 & 28.42 & 48.22 & 50.51 \\
\midrule
Layers 1--6
& \textbf{70.16} & \textbf{63.75} & \textbf{59.25}
& \textbf{46.29} & 60.00
& 47.59 & 53.47 & \textbf{33.30} & 53.74
& \textbf{54.17} \\
Layers 7--12
& 66.49 & 62.25 & 57.88 & 45.66 & 62.67
& 51.32 & 54.07 & 29.20 & 51.53 & 53.45 \\
Layers 13--18
& 67.02 & 62.25 & 56.62 & 45.29 & 62.67
& 52.29 & 54.80 & 30.14 & 54.14 & 53.91 \\
Layers 19--24
& 64.92 & 62.50 & 56.38 & 46.03 & 64.00
& 47.02 & \textbf{55.20} & 32.11 & \textbf{59.01} & 54.13 \\
Layers 25--30
& 69.63 & 59.25 & 53.62 & 46.24 & \textbf{65.00}
& \textbf{53.12} & 53.93 & 28.77 & 54.10 & 53.74 \\
Layers 31--36
& 69.11 & 61.00 & 56.38 & 44.56 & 60.00
& 51.29 & 53.73 & 28.42 & 47.45 & 52.44 \\
\bottomrule
\end{tabular*}
\endgroup
\end{table*}
\section{Diagnostic Protocol Details}
\label{app:v2_diagnostics}
We describe the details of interventions, representation metrics, and readout
conditions used to analyze state formation, evolution, and readout
in Sec.~\ref{sec:analysis}.

\subsection{State Formation}
\label{app:v2_diagnostic_formation}

Our formation analysis examines whether the final decoder
layer is the most effective source of latent states for
subsequent reasoning.
We first test this choice through layer-skipping interventions
and source-range training comparisons.
We then analyze visual alignment across decoder depth
to understand why some sources are more effective than others.
The experimental procedures are detailed below.

\paragraph{Layer skipping.}
\label{app:formation_skipping}
For the inference intervention, we use the frozen
LVR-3B and MONET-7B checkpoints and replace one decoder
block's output hidden state with its input during
latent-state generation.
The block executes before replacement, so its native
key--value (KV) computation and cache updates are preserved.
The remaining blocks process the modified hidden states
normally. The same block is bypassed at all four latent
steps. Answer decoding restores all block transformations
and continues from the resulting cache. Each condition is
compared with the unmodified checkpoint on the same
512 examples.

We measure changes in answer accuracy and extracted
predictions. Let $y_i$ denote the normalized reference
answer for example $i$, and let $\hat y_i^{(\ell)}$ and
$\hat y_i^{\mathrm{base}}$ denote the normalized predictions
with layer $\ell$ skipped and with the unmodified MLLM,
respectively. For $N$ examples,
\begin{equation}
\begin{aligned}
\mathrm{Acc}^{(\ell)}
&=
\frac{100}{N}\sum_{i=1}^{N}
\mathbf{1}\!\left[\hat y_i^{(\ell)}=y_i\right],\\
\Delta\mathrm{Acc}^{(\ell)}
&=
\mathrm{Acc}^{(\ell)}-\mathrm{Acc}^{\mathrm{base}},\\
\Delta\mathrm{Pred}^{(\ell)}
&=
\frac{100}{N}\sum_{i=1}^{N}
\mathbf{1}\!\left[
\hat y_i^{(\ell)}\neq\hat y_i^{\mathrm{base}}
\right].
\end{aligned}
\label{eq:app_formation_skipping}
\end{equation}
Here, $\mathbf{1}[\cdot]$ is the indicator function.
Accuracy changes are reported in percentage points;
prediction changes are percentages of examples. Predictions
are compared after answer extraction and normalization,
so formatting differences alone do not count as changed
answers. Each point in Fig.~\ref{fig:formation}a represents
one skipped decoder layer.

\paragraph{Final task performance.}
\label{app:formation_source_depth}
To determine whether alternative source layers improve
downstream performance after training, we compare six
source-range configurations with the fixed-final baseline
on Qwen2.5-VL-3B.
All runs start from the same instruction-tuned backbone
and follow the official LVR data, objectives, loss weights,
and optimization schedule~\cite{li2025latent}.
The six source ranges are layers 1--6, 7--12, 13--18,
19--24, 25--30, and 31--36.
Each configuration randomly selects a layer within its
assigned range during training and inference.
The selected layer receives visual supervision and supplies the latent states, while the baseline uses layer 36 for both.
All decoder layers remain active, and answer prediction
uses the final decoder output.

We evaluate the final checkpoints with the full image and
question and four latent steps.
Table~\ref{tab:formation_source_depth} extends the main-text
comparison to all nine benchmarks, following the evaluation
protocols in Appendix~\ref{app:v2_evaluation}.
All six source-range configurations outperform the
fixed-final baseline in nine-benchmark mean accuracy,
with gains of 1.93--3.66 percentage points.
These consistent gains show that defaulting to the final
decoder layer limits downstream performance and establish
source-layer selection as a consequential choice in
latent-state formation.

\paragraph{Layerwise visual alignment.}
\label{app:formation_visual_alignment}
We collect representations at the positions that predict
latent inputs, using $N=512$ examples and $K=4$ latent
positions per example. Let
$h_{i,t}^{(\ell)}\in\mathbb{R}^{d}$ denote the
representation for example $i$, latent position $t$,
and decoder layer $\ell$. Non-final layers use raw block
outputs. The final-layer endpoint uses the output after
final RMSNorm, matching the native latent-state interface.

For target-crop cosine similarity, we crop the annotated target region
and encode it independently of the full image. If
$c_{i,1},\ldots,c_{i,M_i}$ are its visual embeddings,
the reference is their mean:
\begin{equation}
\bar c_i=\frac{1}{M_i}\sum_{m=1}^{M_i}c_{i,m}.
\end{equation}
The cosine similarity at each layer and latent position is
\begin{equation}
s_{i,t}^{(\ell)}
=
\frac{
\left\langle h_{i,t}^{(\ell)},\bar c_i\right\rangle
}{
\left\|h_{i,t}^{(\ell)}\right\|_2
\left\|\bar c_i\right\|_2
}.
\label{eq:app_crop_similarity}
\end{equation}
The crop reference remains fixed across layers and latent
positions for a given example. This independently encoded crop
reference differs from the bbox-selected original-image embeddings
used for source selection in Appendix~\ref{app:v2_basis_math}.

For BBox enrichment, let $V_i$ contain all
original-image token positions and $R_i\subseteq V_i$
contain those selected by the annotated bounding box.
We use LVR's official bbox-to-token mapping. Normalized
box coordinates are mapped to the patch grid, with lower
boundaries rounded down and upper boundaries rounded up,
then mapped to the grid after $2\times2$ spatial merging.
The resulting visual-token indices define $R_i$.

Let $a_{i,t,j}^{(\ell,h)}$ denote attention from the
latent-prediction position $t$ to image token $j$ at layer
$\ell$ and head $h$. We average over the $H_\ell$ heads:
\begin{equation}
\bar a_{i,t,j}^{(\ell)}
=
\frac{1}{H_\ell}
\sum_{h=1}^{H_\ell}a_{i,t,j}^{(\ell,h)}.
\end{equation}
BBox enrichment is
\begin{equation}
E_{i,t}^{(\ell)}
=
\frac{
\sum_{j\in R_i}\bar a_{i,t,j}^{(\ell)}
}{
\sum_{j\in V_i}\bar a_{i,t,j}^{(\ell)}
}
\frac{|V_i|}{|R_i|}.
\label{eq:app_attention_enrichment}
\end{equation}
The first ratio is the fraction of image-directed attention
assigned to the region. The second factor corrects for
the region's share of image tokens, not its pixel area.
Thus, $E=1$ indicates attention proportional to token
count, and $E>1$ indicates enrichment above that baseline.

Both metrics are computed separately for each example,
latent position, and layer. We first average over the
four positions within each example, then over examples:
\begin{equation}
\bar m^{(\ell)}
=
\frac{1}{N}\sum_{i=1}^{N}
\left(
\frac{1}{K}\sum_{t=1}^{K}m_{i,t}^{(\ell)}
\right),
\qquad m\in\{s,E\}.
\label{eq:app_alignment_aggregation}
\end{equation}

\subsection{State Evolution}
\label{app:v2_diagnostic_evolution}
Our evolution analysis examines whether effective reasoning
requires states to vary across steps and what their updates
should preserve.
The protocols below cover replay and repetition, analyses
of state geometry and visual focus, and steering experiments
(Fig.~\ref{fig:evolution}).
All experiments use the same 512 examples per model and
retain the original image and question.
Each condition is evaluated in a fresh inference run
with greedy decoding.

\paragraph{Fixed-state replay and repetition.}
\label{app:evolution_replay}
To test whether effective reasoning requires changing
latent inputs across steps, we compare replay of an evolving
trajectory with repetition of a single cached state.
Using the frozen LVR-3B and MONET-7B checkpoints, we generate
latent trajectories from full image--question inputs and
cache the supplied latent states.
For a supplied length $K$, trajectory replay feeds the
cached prefix $(z_1,\ldots,z_K)$ to the reader, while
initial-state repetition supplies $K$ copies of $z_1$.

We also examine whether the choice of repeated state matters
by replacing $z_1$ with a later cached state $z_t$.
For a selected source position $t$, the supplied inputs are
\begin{equation}
\widetilde z_k^{(t)}=z_t,
\qquad k=1,\ldots,K.
\label{eq:app_repeated_feedback}
\end{equation}
Each example uses its own cached state, repeated from
the first latent position.
All conditions retain the original image and question,
boundary markers, and the same number of latent positions.
Each condition rebuilds the reader context and KV cache,
allowing the decoder to continue computing even when
the supplied vectors remain unchanged.
For MONET-7B, these comparisons operate within each
latent segment using its corresponding cached states.
The main-text replay comparisons vary $K$ up to the
default diagnostic budgets of eight states for LVR-3B and ten for
MONET-7B, and the source-state comparisons fix $K$ at
these budgets.

\begin{figure}[t]
    \centering
    \includegraphics[width=0.7\linewidth]{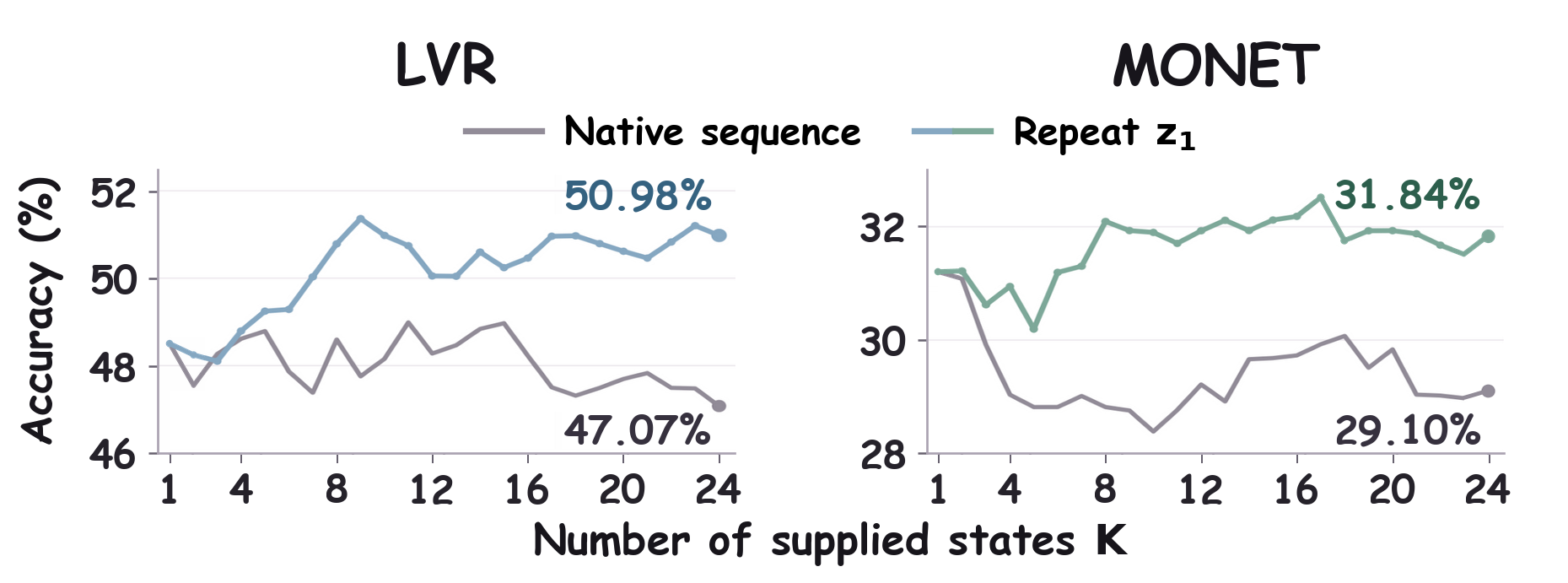}
    \caption{\textbf{Replay and initial-state repetition
beyond the diagnostic budgets.}
    Accuracy of cached trajectory replay and repetition
    of $z_1$ across supplied lengths up to $K=24$,
    evaluated on LVR-3B (left) and MONET-7B (right).
    The original image and question are retained
    in both conditions.}
    \label{fig:extended_replay}
\end{figure}

Further, to examine whether these findings persist beyond
these budgets, we extend the cached trajectories
and supplied lengths to 24 states.
Figure~\ref{fig:extended_replay} compares trajectory replay
with initial-state repetition across these longer horizons.
At $K=24$, repeating $z_1$ achieves 50.98\% accuracy on
LVR-3B, compared with 47.07\% for trajectory replay.
On MONET-7B, the corresponding accuracies are 31.84\%
and 29.10\%.
The advantage of reusing the initial state therefore
persists beyond the diagnostic budgets, showing that useful
latent inputs need not vary for continued computation
to remain effective.

\paragraph{State geometry and visual focus.}
\label{app:evolution_geometry}
\label{app:evolution_attention}
In Fig.~\ref{fig:evolution}b, we examine how state
geometry and visual focus change across reasoning steps
to understand why the initial state is more effective
when reused.
We conduct the following two analyses on the cached native
trajectories, reporting the first eight states for LVR-3B
and ten for MONET-7B.
For MONET-7B, both analyses use the first latent segment
of each example.

We measure \textbf{across-sample dimensionality} using
PCA at each reasoning step.
At position $t$, we collect the corresponding states
from the 512 examples.
For LVR-3B, we L2-normalize each state before centering
the vectors across examples.
For MONET-7B, we center the states without L2 normalization.
Given covariance eigenvalues
$\lambda_{t,1}\geq\cdots\geq\lambda_{t,r}\geq0$,
we define
\begin{equation}
d_{95}(t)
=
\min\left\{
m:
\frac{\sum_{j=1}^{m}\lambda_{t,j}}
     {\sum_{j=1}^{r}\lambda_{t,j}}
\geq0.95
\right\}.
\label{eq:app_evolution_d95}
\end{equation}
This is the smallest number of principal components
explaining at least 95\% of the across-sample variance.
A lower value indicates that variation across examples
is concentrated in fewer directions.

We measure \textbf{visual focus} using final-layer BBox enrichment to track whether successive states
maintain attention to the target region.
Let $a_{i,t,j}^{(L,h)}$ denote attention to original-image
token $j$ for example $i$ at position $t$, from head $h$
of the final decoder layer $L$.
We first average over heads:
\begin{equation}
\bar a_{i,t,j}
=
\frac{1}{H_L}
\sum_{h=1}^{H_L}a_{i,t,j}^{(L,h)}.
\end{equation}
With $V_i$ denoting all original-image positions and
$R_i\subseteq V_i$ the bbox-selected positions,
enrichment is computed as
\begin{equation}
E_{i,t}
=
\frac{
\sum_{j\in R_i}\bar a_{i,t,j}
}{
\sum_{j\in V_i}\bar a_{i,t,j}
}
\frac{|V_i|}{|R_i|}.
\label{eq:app_evolution_attention}
\end{equation}
The bbox mapping and token-count correction follow
Appendix~\ref{app:formation_visual_alignment}.
An enrichment of one indicates attention proportional
to the region's share of image tokens.
At each step, we average enrichment across examples:
\begin{equation}
\bar E_t
=
\frac{1}{N}\sum_{i=1}^{N}E_{i,t},
\qquad N=512.
\label{eq:app_evolution_attention_average}
\end{equation}

\paragraph{Steering of visual grounding and state diversity.}
\label{app:evolution_steering}
To test whether state evolution benefits from visual
grounding or increased cross-step diversity, we compare
three inference-time interventions in
Fig.~\ref{fig:evolution}c.
We use the frozen LVR-3B checkpoint on the same 512 GQA
examples, retaining the full image and question and
four latent steps.
Each intervention modifies the state supplied to the next step, so subsequent states are generated under the
modified history.
We vary the intervention strength
$\beta\in\{0,0.1,\ldots,0.9\}$, with $\beta=0$
corresponding to the unmodified baseline.

Let $h_t$ denote the candidate state before intervention.
We preserve its norm through
$\mathcal N_h(v)=\|h\|_2v/\|v\|_2$.
For \textbf{oracle visual steering}, we select original-image
post-merger tokens using the annotated target boxes.
We construct an orthonormal basis $U_{\mathrm{reg}}$
from at most 128 selected tokens with uniform weights
and a maximum rank of 16, following
Appendix~\ref{app:v2_basis_math}.
The basis remains fixed across steps and intervention
strengths for each example.
Writing $P_x=U_{\mathrm{reg}}U_{\mathrm{reg}}^\top$,
the visual update is
\begin{equation}
z_t^{\mathrm{vis}}
=
\mathcal N_{h_t}
\bigl((1-\beta)h_t+\beta P_xh_t\bigr).
\label{eq:h3_intervention_visual}
\end{equation}
This update attenuates the component outside the
target-region visual subspace.

For the \textbf{random-subspace control}, we sample a
Gaussian matrix and apply QR orthonormalization to obtain
a random basis $Q_x$ with the same actual rank as
$U_{\mathrm{reg}}$.
We replace $P_x$ in Eq.~\ref{eq:h3_intervention_visual}
with $Q_xQ_x^\top$, retaining the same update form,
strength $\beta$, and norm restoration.
Each example uses one random basis, fixed across steps
and intervention strengths.

For \textbf{decorrelation}, we reduce the component
aligned with previously supplied states.
At step $t>1$, let $V_{t-1}$ be an orthonormal basis
for the numerically retained span of the intervened
inputs $\{z_1,\ldots,z_{t-1}\}$.
With $P_{t-1}=V_{t-1}V_{t-1}^\top$, we apply
\begin{equation}
z_t^{\mathrm{div}}
=
\mathcal N_{h_t}
\bigl(h_t-\beta P_{t-1}h_t\bigr),
\qquad t=2,3,4.
\label{eq:h3_intervention_diversity}
\end{equation}
The first input remains unchanged,
$z_1^{\mathrm{div}}=h_1$.
This intervention encourages cross-step diversity
by suppressing components shared with earlier inputs.

We evaluate each intervention through its effects on
answer accuracy and state diversity.
Accuracy uses the normalized exact-match scoring
described in Appendix~\ref{app:v2_diagnostic_scoring}.
For condition $c$ at strength $\beta$, the accuracy
gain is measured against the unmodified four-step
online baseline:
\begin{equation}
\Delta\mathrm{Acc}_{c,\beta}
=
\mathrm{Acc}_{c,\beta}
-
\mathrm{Acc}_{\mathrm{baseline}}.
\end{equation}
Gains are reported in percentage points.

State diversity measures the average cosine
dissimilarity among the states actually supplied
to the model.
For example $i$, let
$\widehat z_{i,t}=z_{i,t}/\|z_{i,t}\|_2$.
We compute
\begin{equation}
D_i
=
\frac{2}{K(K-1)}
\sum_{s<t}
\left(
1-\widehat z_{i,s}^{\top}\widehat z_{i,t}
\right),
\qquad K=4.
\label{eq:h3_intervention_feedback_diversity}
\end{equation}
This averages all six state pairs within each trajectory.
We then average $D_i$ over the 512 examples.
Additional strength-matched controls are described in
Appendix~\ref{app:steering_matched_controls}.



\begin{table*}[t]
\centering
\caption{\textbf{Visual supervision and inference state source.}
Accuracy (\%) on LVR-3B.
Rows sharing a supervision layer use identical checkpoint
weights; only the inference state source changes.
The L36/L36 row reproduces the main-table baseline.
Avg.\ is the unweighted mean across nine benchmarks.
Bold marks the highest score within each supervision group,
including ties.}
\label{tab:formation_crossed_ablation}

\begingroup
\fontsize{8.5}{9.5}\selectfont
\setlength{\tabcolsep}{2.5pt}
\renewcommand{\arraystretch}{1.08}

\begin{tabular*}{\textwidth}{
    @{\extracolsep{\fill}}cc*{10}{r}@{}
}
\toprule
\shortstack{Supervision\\layer}
& \shortstack{Source\\layer}
& V$^{\star}$ & HR4K & HR8K & BLINK & MMVP
& MME-RW & MMStar & VisPuz & CVBench & Avg. \\
\midrule
L36 & L36
& \textbf{66.49} & 56.38 & 52.75
& \textbf{43.66} & \textbf{61.33}
& 47.65 & 49.67 & \textbf{28.42} & 48.22 & 50.51 \\
L36 & L3
& 65.97 & \textbf{61.62} & \textbf{55.62}
& \textbf{43.66} & 60.00
& \textbf{49.57} & \textbf{50.93} & 28.00
& \textbf{52.19} & \textbf{51.95} \\
\midrule
L3 & L36
& 68.59 & 57.50 & 53.75 & 44.77 & 59.33
& 51.00 & 49.60 & 26.46 & 39.30 & 50.03 \\
L3 & L3
& \textbf{70.16} & \textbf{63.87} & \textbf{57.25}
& \textbf{45.87} & \textbf{61.67}
& \textbf{51.31} & \textbf{52.27} & \textbf{28.34}
& \textbf{45.66} & \textbf{52.93} \\
\bottomrule
\end{tabular*}
\endgroup
\end{table*}

\subsection{State Readout}
\label{app:v2_diagnostic_readout}

Our readout analysis examines whether the model effectively
uses the visual and task information carried by latent states.
We first withhold the image or question to examine native
readout, then test whether adapting the readout of fixed
states improves the use of retained task information.
The corresponding experimental procedures are described below.

\paragraph{Context removal and content references.}
\label{app:readout_context_removal}
To examine what the reader can recover from latent states,
we separate state generation from answer prediction.
LVR-3B and MONET-7B each generate ten latent states from
the full image--question input for every diagnostic example.
MONET-7B also caches its intermediate text and retains
the interleaved order during readout.
We then supply this fixed content under three contexts:
the full image and question, the question with the image
withheld, or the image with the question withheld.
All conditions share the same generic answer instruction.
Each reader context and KV cache is rebuilt independently
to exclude the withheld input.

We additionally provide explicit references for visual
and task information.
The \textbf{visual references} consist of spatially ordered,
unpooled BBox-selected tokens from the original image encoding,
using the mapping in
Appendix~\ref{app:formation_visual_alignment}.
The \textbf{task-plan references} consist of four-step plans
describing what to locate, inspect, compare, or verify,
without reporting observations, deductions, or answers.
Their generation procedure and prompt are provided in
Appendix~\ref{app:task_plan_generation}.

For LVR-3B, either reference replaces the latent sequence.
For MONET-7B, visual references replace the latent states
while retaining the cached text; task plans replace both
the text and latent states.
For each reference condition $c$, the reported accuracy
is averaged across the two readers:
\begin{equation}
\overline{\mathrm{Acc}}_c
=
\tfrac12\bigl(
\mathrm{Acc}_{\mathrm{LVR},c}
+
\mathrm{Acc}_{\mathrm{MONET},c}
\bigr).
\label{eq:app_reference_average}
\end{equation}

\paragraph{Adapter-based task-readout diagnostic.}
\label{app:trained_state_readout}
To test whether task information is retained but underused,
we adapt the readout interface while keeping both the source
states and reader backbone fixed.
The frozen LVR-3B checkpoint generates four-state caches
from full image--question inputs.
A shared residual MLP adapter transforms each cached state
before readout.
Only the adapter is trained; the source caches are not
regenerated after adapter updates.

We use 512 training and 128 validation image--question
pairs from the GQA validation data in Visual-CoT.
The original GQA512 diagnostic set is held out for testing,
with no image overlap among the three splits.

For a cached state $z\in\mathbb{R}^{d}$, with $d=2048$,
the adapter applies
\begin{align}
r(z)
&=\left(\frac{\|z\|_2^2}{d}+10^{-6}\right)^{1/2},\\
\mathcal{A}_{\phi}(z)
&=z+gW_{\mathrm{up}}\,
\operatorname{GELU}\!\left(
W_{\mathrm{down}}\frac{z}{r(z)}
\right).
\end{align}
Here, $W_{\mathrm{down}}\in\mathbb{R}^{64\times2048}$
and $W_{\mathrm{up}}\in\mathbb{R}^{2048\times64}$
are bias-free linear layers, and $g$ is a trainable scalar.
The adapter is shared across all examples and latent
positions and contains 262,145 trainable parameters.
We initialize $W_{\mathrm{down}}$ from
$\mathcal{N}(0,0.02^2)$, set $W_{\mathrm{up}}=0$,
and initialize $g=0.1$, giving an initially identity mapping.
The adapter operates in FP32 and casts its outputs back
to the cached-state dtype before readout.

The reader receives the image, four adapted states,
and the generic instruction $G$ specified in
Appendix~\ref{app:v2_combo}, while the original question
is withheld.
During both training and evaluation, the reader context
and KV cache are rebuilt independently, without reusing
the generation cache containing the original question.
Only answer tokens receive teacher-forced cross-entropy
supervision, averaged over answer tokens within each example.

To discourage changes to the original states, we add
a residual penalty.
For example $i$, let $z_{i,k}$ be the original cached state
and $\widehat z_{i,k}$ its adapted value after the dtype cast.
With $K=4$ latent positions and effective batch size $B$,
we optimize
\begin{align}
\mathcal{R}_i
&=\frac{1}{Kd}\sum_{k=1}^{K}
\frac{\|\widehat z_{i,k}-z_{i,k}\|_2^2}
     {r(z_{i,k})^2},\\
\mathcal{L}
&=\frac{1}{B}\sum_{i=1}^{B}
\left(\mathcal{L}_{\mathrm{ans},i}
+10^{-4}\mathcal{R}_i\right),
\end{align}
where $\mathcal{L}_{\mathrm{ans},i}$ is the answer-token
cross-entropy for example $i$.
The penalty is evaluated in FP32 and measures the mean
squared RMS-normalized state change.

We use AdamW with a constant learning rate of $10^{-3}$,
zero weight decay, and gradient-norm clipping at $1.0$.
The microbatch size is one, with gradients averaged
over eight examples per update, giving $B=8$.
Adapters are trained for eight epochs, totaling
512 optimizer updates, using seeds 42, 43, and 44.
All evaluations use the final eighth-epoch weights.

At evaluation, we compare three conditions:
\textbf{correct states} generated for the original
image--question pair,
\textbf{wrong-question states} generated for another
question about the same image, and
\textbf{no latent states}.
All predictions are scored against the answer to the
original question.
The baseline reads the original caches, while the adapter
transforms those same caches before readout.
We use greedy decoding and normalized exact-match scoring,
and average adapter results across the three training seeds.
The shuffled-cache control shares the same architecture
and training settings; its protocol and the complete
results are provided in
Appendix~\ref{app:adapter_readout_controls}.

\section{Additional Diagnostic Controls}
\label{app:additional_diagnostic_controls}
The following controls test alternative explanations
for the formation, evolution, and readout findings
in Sec.~\ref{sec:analysis}.
\subsection{Visual Supervision and State Source}
\label{app:formation_crossed_ablation}

To test whether the benefit of earlier-layer state sources can be explained solely by changing visual supervision,
we vary the supervision layer and inference state source independently.
We use the standard LVR-3B checkpoint supervised at
layer 36 and train an additional checkpoint with visual
MSE supervision at layer 3 for 5,000 updates following
the official LVR procedure.
Each checkpoint is evaluated using either layer 3
or layer 36 as the fixed state source. During training, each checkpoint extracts latent states
from the layer receiving visual supervision.  Within each checkpoint, only the state source changes;
the weights and supervision remain fixed.
All conditions retain the full image and question,
use four latent steps, and keep all decoder layers active.

Table~\ref{tab:formation_crossed_ablation} shows that
switching the state source from layer 36 to layer 3 improves
mean accuracy under both supervision configurations.
For the layer-36-supervised checkpoint, the gain is
1.44 percentage points.
For the layer-3-supervised checkpoint, mean accuracy
increases from 50.03\% to 52.93\%.
The benefit of earlier-layer state sources therefore persists
when visual supervision is held fixed, ruling out
supervision relocation as its sole explanation.

\begin{table*}[t]
\centering
\caption{\textbf{Final-layer pre-RMSNorm comparison.}
Accuracy (\%) on Qwen2.5-VL-3B.
Avg.\ is the unweighted nine-benchmark mean.}
\label{tab:formation_pre_rmsnorm}
\begingroup
\small
\setlength{\tabcolsep}{2pt}
\renewcommand{\arraystretch}{1.05}
\begin{tabular*}{\textwidth}{@{\extracolsep{\fill}}l*{10}{r}@{}}
\toprule
Configuration & V$^\star$ & HR4K & HR8K & BLINK & MMVP
& MME-RW & MMStar & VisPuz & CVBench & Avg. \\
\midrule
LVR
& 66.49 & 56.38 & 52.75 & 43.66 & 61.33
& 47.65 & 49.67 & 28.42 & 48.22 & 50.51 \\
Pre-RMSNorm
& 67.54 & 60.00 & 54.37 & 43.71 & 60.67
& 46.96 & 51.47 & 26.03 & 44.41 & 50.57 \\
\bottomrule
\end{tabular*}
\endgroup
\end{table*}

\subsection{Final-Layer Normalization}
\label{app:formation_normalization}

To test whether the formation results can be explained by
normalization at the final-layer state interface, we compare
the standard LVR-3B baseline with a separately trained
pre-RMSNorm variant.
The variant uses the final decoder layer's output before
final RMSNorm as the latent state, retaining the final layer as
the state source.

As shown in Table~\ref{tab:formation_pre_rmsnorm},
the pre-RMSNorm variant achieves a nine-benchmark mean
accuracy of 50.57\%, compared with 50.51\% for the baseline.
Removing final-layer normalization leaves mean performance
nearly unchanged and does not reproduce the gains from
alternative source layers.
This control indicates that the formation results
cannot be explained simply by the normalization difference.

\subsection{Controls for Fixed-State Repetition}
\label{app:fixed_feedback_controls}

To examine whether the benefit of repetition depends on
the supplied state, we compare five supplied-state conditions
using the frozen LVR-3B checkpoint on the same
512 GQA examples.
For each supplied length $K=1,\ldots,24$, we evaluate
the cached trajectory prefix, repetition of the example's
own first state, repetition of another image's first state,
a shared fixed vector, and zero vectors.
All conditions retain the original image and question,
boundary markers, and $K$ latent positions. The reader continues computing at every position,
even when the supplied vectors remain unchanged.

For the other-image control, examples are randomly paired
without self-matches or answer-based selection.
Each donor state is rescaled to match the recipient's
first-state norm, with the pairing fixed across lengths.
The shared-vector control uses one Gaussian vector for
all examples and lengths, scaled by the median first-state
norm from training images disjoint from the test set.
The zero control retains all latent positions and thus
differs from removing the latent span.
Each condition uses a fresh reader context and KV cache,
greedy decoding with a 32-token answer budget, and
normalized exact-match scoring.

\begin{figure}[t]
    \centering
    \includegraphics[width=0.85\linewidth]
        {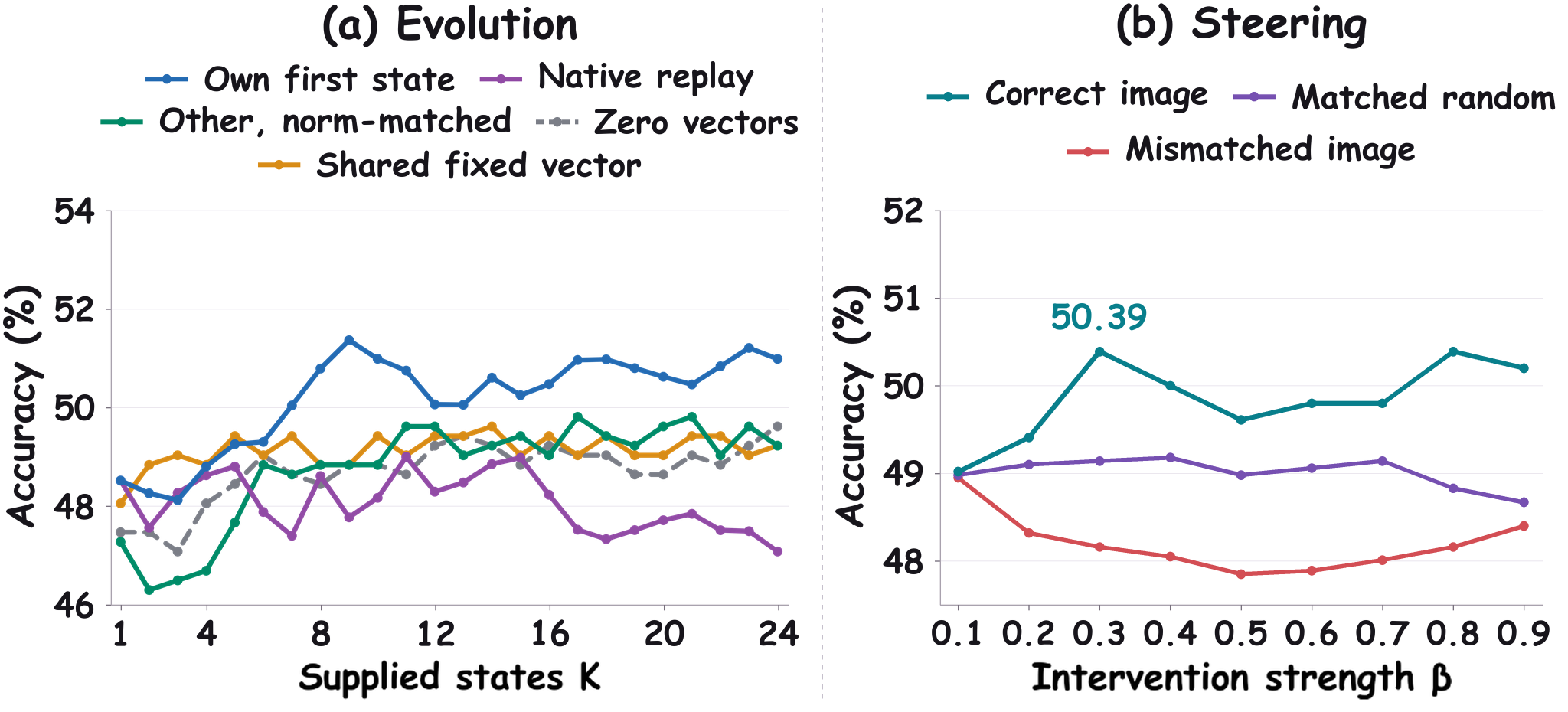}
    \caption{\textbf{Controls for state repetition and visual steering.}
    Accuracy (\%) on GQA512 using frozen LVR-3B,
    with the original image and question retained.
    \textbf{(a)} Native replay and repeated-state controls
    across supplied lengths.
    \textbf{(b)} Correct-image steering versus strength-matched
    random-direction and mismatched-image controls with four
    latent steps. Control curves average five assignments.}
    \label{fig:evolution_steering_controls}
    \label{fig:evolution_feedback_controls}
    \label{fig:steering_matched_controls}
\end{figure}

As shown in Fig.~\ref{fig:evolution_feedback_controls}a,
repeating the example's own first state generally
outperforms the fixed and mismatched controls.
At $K=8$, it reaches 50.78\%, compared with 48.83\%
for both the norm-matched other-image state and the shared
vector, and 48.44\% for zeros.
Fixed and zero states can also outperform native replay
at longer lengths, so an advantage over the original
trajectory is not unique to sample-matched content.
Nevertheless, the additional gains from own-state repetition
show that the choice of repeated state still matters,
even when the number and format of supplied states remain unchanged.

\subsection{Strength-Matched Controls for Visual Steering}
\label{app:steering_matched_controls}

We test whether the advantage of visual steering depends on
image-matched information rather than intervention strength alone.
Using the frozen LVR-3B checkpoint and online intervention
protocol in Appendix~\ref{app:evolution_steering}, we compare
correct-image visual guidance with two controls:
a random direction and visual guidance derived from a
mismatched image.
We quantify realized intervention strength by the relative
displacement $\delta_t=\|z_t-h_t\|_2/\|h_t\|_2$, where
$h_t$ is the candidate state and $z_t$ is the final intervened
state. For each example, latent step, and tested $\beta$, both controls
match the relative displacement $\delta_t$ measured on the
correct-image trajectory, each preserving the norm of its own
candidate state ($\|z_t\|_2=\|h_t\|_2$).
All conditions retain the full image and question,
four latent steps, and the same decoding and scoring settings.
No additional training is performed.

We evaluate $\beta\in\{0.1,0.2,\ldots,0.9\}$ on the same
512 GQA examples using control-assignment seeds
$s\in\{42,43,44,45,46\}$.
These seeds vary the random-direction and mismatched-image
assignments while keeping the model checkpoint and evaluation
examples fixed.
For each control $c$ and strength $\beta$, we report
\begin{equation}
\overline{A}_{c}(\beta)
=\frac{1}{5}\sum_{s\in\{42,43,44,45,46\}}
\frac{100\,m_{c,s}(\beta)}{512},
\end{equation}
where $m_{c,s}(\beta)$ is the number of correct answers.
All outputs remain in the denominator, including truncated
outputs, and rounding is applied only after averaging.
Correct-image guidance does not depend on the control
assignment and is evaluated once.

As shown in Fig.~\ref{fig:steering_matched_controls}b,
correct-image guidance achieves higher accuracy than both
control means across the tested strengths.
This comparison supports a contribution from image-matched
visual information beyond perturbation strength alone.

\subsection{Controls for Adapter-Based Readout}
\label{app:adapter_readout_controls}

To examine whether the adapter benefits from correctly
paired task information, we compare paired training
with a shuffled-cache control.
Paired training preserves the correspondence between
each latent-state cache and its target image and answer.
Shuffled training permutes the caches across training
examples while leaving the target images and answers
unchanged.
For each of seeds 42, 43, and 44, the two adapters share
the same initialization, example order, and training budget.
Both conditions use correctly paired validation caches,
and all evaluations use the final eighth-epoch weights.

Table~\ref{tab:adapter_readout} reports the complete results.
With the question withheld, paired training achieves
26.17\% accuracy with correct states, compared with
18.03\% for shuffled training.
The correct-minus-wrong-question advantage is
6.58 percentage points for the paired adapter and
1.24 points for the shuffled adapter.
Preserving the state--task correspondence during training
therefore strengthens the use of task-matched states.

To test whether adapter gains extend to full-context
answering, we evaluate the same adapters with the original
image and question both available, without additional
training.
The paired adapter raises correct-state accuracy from
46.88\% to 51.43\%.
It also improves accuracy with wrong-question states,
which reach 50.78\%.
The adaptation thus benefits full-context answering
as well as question-withheld readout.

\begin{table}[t]
\centering
\caption{\textbf{Controls for adapter-based readout.}
Accuracy (\%) on GQA512 using four cached states.
Adapter results average training seeds 42, 43, and 44
at epoch eight; LVR-3B is the frozen baseline.
Full-context evaluation reuses the same trained adapters.
Wrong Q denotes states generated for another question
about the same image.
Bold marks the highest correct-state accuracy in each context.}
\label{tab:adapter_readout}

\begingroup
\fontsize{8.5}{9.5}\selectfont
\setlength{\tabcolsep}{3pt}
\begin{tabular*}{\linewidth}{
    @{\extracolsep{\fill}}lrrr@{}
}
\toprule
Reader & Correct & Wrong Q & No latent \\
\midrule
\multicolumn{4}{l}{\textit{Question withheld: $I+G$}} \\
LVR-3B
& 17.38 & 13.09 & 0.59 \\
Paired adapter
& \textbf{26.17} & 19.60 & 0.59 \\
Shuffled adapter
& 18.03 & 16.80 & 0.59 \\
\midrule
\multicolumn{4}{l}{\textit{Full context: $I+Q$}} \\
LVR-3B
& 46.88 & 45.51 & 47.66 \\
Paired adapter
& \textbf{51.43} & 50.78 & 47.66 \\
Shuffled adapter
& 49.48 & 49.48 & 47.66 \\
\bottomrule
\end{tabular*}
\endgroup
\end{table}

\section{Method Implementation Details}
\label{app:method_implementation}

This section describes how LIFE-VLR implements visually aligned state formation, visually grounded state evolution, and explicit state readout training within LVR and MONET.
Training data, optimization settings, and budgets are provided in
Appendix~\ref{app:v2_implementation}.

\subsection{Source Selection and Visual Bases}
\label{app:v2_basis_math}

Source selection determines which decoder representation supplies the latent states. The visual bases define the directions used for grounding supervision and state steering.

\paragraph{Cosine-based source selection.}
Decoder-block indices are one-based and refer to raw
outputs before final RMSNorm. All non-final blocks are
candidates. A full image--question pass ending at the
latent-start marker produces the first-state prediction
at each block. We select the block with the highest
cosine similarity to the mean target-region embedding,
as in Eq.~\ref{eq:v2_formation}.

For LVR, scores are averaged over the local microbatch
to select one shared source. For MONET, selection is
performed separately for each sample and latent segment.
The decision is treated as constant during backpropagation
and remains fixed within the trajectory or segment.
The selection prompt does not include the answer.

\paragraph{Visual references.}
For LVR, target features are bbox-selected post-merger
embeddings from the original image. We merge and
deduplicate token indices across boxes. If their union
exceeds the visual-token budget, deterministic, box-balanced
sampling selects the retained tokens. The mean of these raw
features is the cosine-selection target. For MONET,
the corresponding auxiliary crop or visual-operation
image supplies the target features. All targets are
detached and are not supplied as additional answer-reader
inputs in the final student stage.

Target features define the grounding basis
$U_{\mathrm{reg}}$. A separate basis,
$U_{\mathrm{attn}}$, guides state steering. It uses
original-image tokens selected by first-state
attention associated with the chosen source block
(Appendix~\ref{app:v2_inference}). Attention is
reconstructed from native rotary-positioned queries
and cached keys, normalized over valid keys, and
averaged over heads. We retain the 128 original-image tokens with the highest
head-averaged attention weights, or all original-image
tokens if fewer than 128 are available. Only the retained original-image
weights enter the steering basis; auxiliary region
images are excluded.

\paragraph{Basis construction and numerical treatment.}
For either token set, we normalize each feature to a
direction $x_j$ and its nonnegative weight to $w_j$,
where $\sum_j w_j=1$. Region-token weights are uniform;
steering-token weights are the selected attention weights.
We preserve the nonzero mean direction
$m=\sum_j w_jx_j$ and obtain the remaining directions
by weighted PCA of mean-centered residuals orthogonal
to $m$. Reduced QR decomposition produces an orthonormal
basis capped at the configured rank.

These operations use float32 and discard numerically
empty directions. Invalid or zero-rank targets are
rejected. Visual features, bases, and source decisions
remain fixed within a trajectory and are treated as
constants during backpropagation. Token and rank budgets are
specified in Appendix~\ref{app:v2_implementation}.
Steering attenuates the component orthogonal to $U_{\mathrm{attn}}$ and
restores the raw prediction's norm, following
Eq.~\ref{eq:v2_feedback}.

\subsection{LVR Training Procedure}
\label{app:v2_combo}
\label{app:v2_algorithm}

LVR training combines state formation under the full image--question
context with answer prediction under the scheduled reader context.
State generation and readout share the language-model parameters.
Each update proceeds as follows:
\begin{enumerate}
\setlength{\itemsep}{2pt}
\item Encode the original image, construct detached region targets,
      and select the source block from the full-context prompt pass.
\item Generate the latent sequence without retaining the generation graph, and save the steered states $Z=(z'_1,\ldots,z'_K)$.
\item Replay the full prompt and saved latent prefix with gradients
      to recompute the raw latent predictions $z_t$ used in the
      grounding loss.
\item Read the saved inputs under the scheduled context and optimize
      answer-token CE. Full-context training shares one replay
      between the two losses. Restricted-context training uses
      a separate full-context formation replay, without the answer,
      for grounding.
\end{enumerate}

The grounding loss is averaged over latent positions.
Both losses update the shared model parameters, while gradients
do not pass through the cached rollout. Subsequent batches generate
new states with the updated parameters.

The reader curriculum progresses from full-context readout to
readout without the image and then readout without the question.
Withholding the image deletes the visual span and its associated inputs.
Withholding the question retains the pre-decoder image features and replaces
the original question, answer options, and system text with
\emph{Answer the original question using the latent information.}
For each context, we rebuild the input sequence, attention mask,
positional indices, and KV cache to exclude the withheld information.
Only answer tokens receive CE supervision.

\subsection{MONET Training Procedure}
\label{app:v2_monet}
\label{sec:v2_monet}

MONET applies the same interventions within its three-stage
distillation pipeline. The stages determine how explicit visual
observations are converted into latent segments and how those
segments are used during answer prediction.

\paragraph{Stage construction.}
Stage~1 learns the multimodal demonstrations through supervised
text prediction. Within each backbone, the baseline and +Ours
share the resulting Stage~1 checkpoint and its observation-teacher
cache. Stage~2 initializes from this checkpoint and distills
from the shared observation cache.

Each branch then uses its own Stage~2 model to construct a
latent-teacher cache for Stage~3. The Stage~3 student also
initializes from the shared Stage~1 checkpoint and learns from
the branch-specific Stage~2 cache. Stage~2 therefore supplies
Stage~3 with distillation targets. Stage~3 replaces auxiliary
images with latent segments while preserving their interleaved
order with the reasoning text.

\paragraph{Interventions and objectives.}
Our source selection, state steering, and grounding objective
apply in both Stages~2 and~3. Source selection is performed per
sample and latent segment, with the source and both visual bases
fixed within that segment. Native detach boundaries between
successive latent inputs are preserved.

Stage~3 additionally uses the reader curriculum described above.
State formation retains the full context for native distillation
and grounding, while the reader computes text CE under the
scheduled context. Grounding loss is added once per update.
Native text and representation-distillation losses,
teacher--student token and layer matching, loss normalization,
and stage-specific gradient routing follow the official MONET
implementation~\cite{wang2025monet}.

\paragraph{Teacher caches and input construction.}
Teacher caches are generated without gradients. Observation caches
are shared within each backbone, while latent caches are specific
to the baseline or +Ours branch.

In Stage~2, native helper-image visibility rules separate auxiliary
visual inputs used for state formation from the text reader.
Stage~3 removes auxiliary images from both student and reader
inputs; their separately encoded features remain as detached
grounding targets. Readers retain the interleaved text and latent
order and reconstruct the retained prompt and positions.
Teacher tensors and auxiliary-image features serve as training
targets and are not inserted into the answer reader.

\subsection{Inference and Backbone Adaptation}
\label{app:v2_inference}

At inference, we use the training-selected source layer and
construct the steering basis from the original image.
The following rules specify source selection and adaptation
to the native backbone interfaces.

\paragraph{Source selection and steering.}
We aggregate source-selection counts across training devices
and save the histogram with the checkpoint. Inference uses
the global modal block for all examples, with ties resolved
in favor of the shallower block.
LVR counts cover the complete reader curriculum.
MONET uses Stage~2 counts when constructing its latent-teacher
cache and Stage~3 counts for final inference.

The steering basis uses original-image embeddings selected
by the model's image--question-conditioned attention.
No target-region annotations are required at test time.
LVR starts a fresh full-context answer reader; MONET retains
its interleaved text and latent generation.
All decoder blocks execute, with the selected block supplying the latent states.

\paragraph{Backbone-specific interfaces.}
All backbones retain their native visual features, positional
encodings, and caches. Qwen3-VL keeps DeepStack features when
the image is available and removes them in the image-withheld
reader.

Qwen3.5 retains its hybrid attention and recurrent cache.
When the selected source is a linear-attention block, the steering-basis attention weights come from the nearest preceding full-attention block,
or the first following full-attention block if none precedes it.
The latent prediction still comes from the selected source.
In MONET's Qwen3.5 adaptation, the text reader removes
auxiliary-image spans and reconstructs native positions and
the recurrent cache. State formation retains the required
visual context.
\begin{table}[!t]
\centering
\caption{\textbf{LVR supervised-training configuration.}
3B, 8B, and 27B denote Qwen2.5-VL-3B, Qwen3-VL-8B, and
Qwen3.5-27B. Batch sizes count image--question examples except
for the explicitly marked packed input. All runs use eight GPUs.}
\label{tab:v2_training_settings}
\small
\setlength{\tabcolsep}{3pt}
\renewcommand{\arraystretch}{1.08}
\begin{tabular*}{\columnwidth}{@{\extracolsep{\fill}}lccc@{}}
\toprule
Setting & 3B & 8B & 27B \\
\midrule
Optimizer updates & 5,000 & 5,000 & 2,500 \\
Per-GPU microbatch & 4 & 4 & 1 pack \\
Gradient accumulation & 1 & 1 & 1 \\
Effective example batch & 32 & 32 & $\leq32$ \\
Packing budget & 16,384 & 16,384 & 16,384 \\
Examples per packed item & 1 & 1 & $\leq4$ \\
DeepSpeed ZeRO stage & 2 & 2 & 3 \\
\midrule
\multicolumn{4}{l}{\textit{Reader curriculum: optimizer updates}} \\
Full context & 1,000 & 1,000 & 500 \\
Image withheld & 2,000 & 2,000 & 1,000 \\
Question withheld & 2,000 & 2,000 & 1,000 \\
\midrule
\multicolumn{4}{l}{\textit{Latent generation}} \\
Latent steps $K$ & 4 & 4 & 4 \\
Candidate source blocks & 1--35 & 1--35 & 1--63 \\
Maximum basis rank & 16 & 16 & 16 \\
Tokens per visual basis & $\leq128$ & $\leq128$ & $\leq128$ \\
Steering strength $\beta$ & 0.2 & 0.2 & 0.2 \\
Grounding weight $\lambda$ & 0.1 & 0.1 & 0.1 \\
\bottomrule
\end{tabular*}
\end{table}

\subsection{Reinforcement-Learning Implementation}
\label{app:rl_implementation}

Continued RL follows the native policy-optimization procedure
of each framework. The baseline and +Ours use the same recipe
within each framework, while retaining their respective source-selection and steering rules.

\paragraph{LVR: reward and gradient scope.}
The reward is the equally weighted sum of binary answer
correctness and valid output format.
Correctness uses symbolic verification followed by exact
matching of the extracted answer. Valid format requires
latent-start and latent-end markers and a final answer
enclosed in answer tags.

Within each group, rewards are centered and divided by their
sample standard deviation plus $10^{-4}$.
The clipped token loss is averaged over valid response tokens
within each microbatch. Truncated completions retain their
valid-token mask.

Only sampled text tokens contribute policy likelihoods;
continuous latent inputs and control positions are excluded.
Latent states are generated without gradients and supplied to a
differentiable answer replay. RL updates the shared language
parameters through answer likelihoods.
The saved source block remains fixed, and +Ours retains
its image-derived state steering.
Supervised losses and the restricted-context reader curriculum
are not used during RL.

\paragraph{MONET: joint policy optimization.}
MONET follows its official VLPO implementation for continuous-latent
and text policy objectives, Gaussian likelihoods, rewards, and
answer-judge handling~\cite{wang2025monet}.
Both branches retain their respective source and steering rules.
Optimization, sampling, and data settings for both frameworks
are provided in Appendix~\ref{app:v2_rl_training}.

\section{Training Details}
\label{app:v2_implementation}

We report the data, initialization, and training settings
used for supervised training and RL.
Supervised experiments use Qwen2.5-VL-3B, Qwen3-VL-8B,
and Qwen3.5-27B, while RL experiments use Qwen2.5-VL-3B.
Within each framework and backbone, the baseline and +Ours are
compared under matched training conditions and the same evaluation
protocol. For supervised evaluation, we use the final LVR checkpoint
and the final Stage~3 checkpoint for MONET. MONET teacher caches
are built from the final checkpoints of Stages~1 and~2. Details are provided below.

\subsection{LVR Supervised Training}
\label{app:v2_training}

\paragraph{Data and initialization.}
We train on Visual-CoT in the LVR question--answer format,
including document-oriented examples.
Each example contains the original image, question,
answer, and target bounding boxes.
The image and question provide the state-generation input,
and bbox-selected visual features provide supervision.
The baseline and +Ours initialize independently from
the same released backbone: Qwen2.5-VL-3B-Instruct,
Qwen3-VL-8B-Instruct, or Qwen3.5-27B.

\paragraph{Trainable parameters.}
We update the language backbone, token embeddings,
and output head.
The visual encoder and merger remain frozen, including
Qwen3-VL's DeepStack mergers.
No additional trainable adapter or prediction head is used.
\begin{table}[!t]
\centering
\caption{\textbf{MONET supervised-training configurations.}
Durations specify the training budgets for each backbone.
Microbatch/accumulation entries are per GPU; eight GPUs give
an effective batch of 128 throughout. Image budgets count
post-merger tokens in the training inputs.}
\label{tab:v2_monet_training}
\footnotesize
\setlength{\tabcolsep}{3pt}
\renewcommand{\arraystretch}{1.08}
\begin{tabular*}{\columnwidth}{@{\extracolsep{\fill}}lccc@{}}
\toprule
Setting & Stage 1 & Stage 2 & Stage 3 \\
\midrule
Initialization & Backbone & Stage 1 & Stage 1 \\
\midrule
\multicolumn{4}{l}{\textit{Training duration and effective batch}} \\
3B/8B duration & 4 epochs & 1,000 steps & 1,000 steps \\
27B duration & 1 epoch & 500 steps & 500 steps \\
3B microbatch/accum. & $8/2$ & $4/4$ & $4/4$ \\
8B microbatch/accum. & $4/4$ & $2/8$ & $2/8$ \\
27B microbatch/accum. & $1/16$ & $1/16$ & $1/16$ \\
\midrule
\multicolumn{4}{l}{\textit{Shared latent and supervision settings}} \\
Latent positions/segment & -- & 8 & 8 \\
Observation CE weight & 2 & 4 & 4 \\
Native alignment coefficient & -- & 2 & 2 \\
Latent-only multiplier & -- & 2 & -- \\
\midrule
\multicolumn{4}{l}{\textit{Image-token caps: total / per image}} \\
3B/8B & $2{,}000/1{,}280$ & $1{,}500/1{,}280$ & $2{,}000/2{,}000$ \\
27B & $2{,}000/1{,}280$ & $768/768$ & $1{,}024/1{,}024$ \\
\bottomrule
\end{tabular*}
\end{table}
\paragraph{Optimization and training budget.}
We use AdamW with peak learning rate $10^{-5}$,
$(\beta_1,\beta_2)=(0.9,0.999)$, $\epsilon=10^{-8}$,
weight decay $0.1$, and gradient-norm clipping at $1.0$.
The learning rate increases linearly over the first 3\%
of updates, then follows cosine decay.
Training uses eight GPUs, bfloat16, gradient checkpointing,
and seed~42 for model and data randomness.

Table~\ref{tab:v2_training_settings} lists the update
budgets, batch sizes, and method hyperparameters.
The baseline's visual-MSE coefficient is $0.1$.
Our grounding objective follows Eq.~\ref{eq:v2_grounding},
with its weight and the steering strength given in the table.
Only the reader context changes across curriculum stages;
the optimizer and learning-rate schedule continue without
restarting, and state generation retains the full image
and question throughout.

\paragraph{Input processing.}
Images retain their aspect ratio under the native processor,
with a budget of 128--5,120 post-merger tokens.
The spatial stride per merged token is 28 pixels for
Qwen2.5-VL and 32 for Qwen3-VL and Qwen3.5.
The corresponding pixel bounds are 100,352--4,014,080
and 131,072--5,242,880, respectively.

The packing budget is 16,384 tokens.
For 3B and 8B, each packed item contains one example,
and each optimizer update uses 32 examples.
For 27B, each GPU processes one packed item containing
at most four examples.
Its effective example batch varies with sequence length
and is bounded by 32.


\subsection{MONET Supervised Training}
\label{app:monet_training_configuration}

\paragraph{Data and initialization.}
MONET uses Monet-SFT-125K, comprising Visual-CoT, CogCoM,
ReFocus, and the counting, visual-search, and geometry
subsets of Zebra-CoT.
Training retains the interleaved reasoning text,
observations, and latent segments.
Within each backbone, the baseline and +Ours share
one Stage~1 checkpoint.
The stage initializations and training budgets are listed
in Table~\ref{tab:v2_monet_training}; the three-stage
procedure is described in Appendix~\ref{app:v2_monet}.

\paragraph{Trainable parameters.}
The language backbone, token embeddings, and output head
are trainable in all stages.
The visual encoder and merger remain frozen.

\paragraph{Optimization and training budget.}
All stages use AdamW with learning rate $10^{-5}$,
$(\beta_1,\beta_2)=(0.9,0.999)$, $\epsilon=10^{-8}$,
weight decay $0.01$, and gradient clipping at $1.0$.
Each stage starts a fresh optimizer and linear
learning-rate schedule with 10 warmup updates.
Training uses eight GPUs, seed~42, bfloat16,
and non-reentrant gradient checkpointing.
Durations and effective batch sizes are matched between
the baseline and +Ours within each backbone.

The 3B and 8B runs use ZeRO-2.
For 27B, Stage~1 uses ZeRO-3 with parameter and optimizer
CPU offload; Stages~2 and~3 use ZeRO-2 with CPU
optimizer offload.

Candidate source blocks are 1--35 for 3B and 8B,
and 1--63 for 27B.
Each visual basis uses at most 128 tokens and rank
at most 16.
Steering strength is $\beta=0.2$, and grounding weight
is $\lambda=0.1$.
The Stage~3 reader curriculum allocates 20\% of updates
to full-context readout, 40\% to readout without the image,
and 40\% to readout with the image and generic instruction
but without the question.
The resulting schedules are 200/400/400 updates for
3B and 8B, and 100/200/200 for 27B.

\paragraph{Input processing.}
Images retain their aspect ratio under the total and
per-image caps in Table~\ref{tab:v2_monet_training}.
The merged-token stride is 28 pixels for 3B and 32
for 8B and 27B.
The Stage~2 total cap includes the original and auxiliary
images.
For 27B, the image-token budgets are 768 in Stage~2
and 1,024 in Stage~3.


\begin{table}[!t]
\centering
\caption{\textbf{RL configuration on Qwen2.5-VL-3B.}
Batch sizes count completions, including the eight samples
per prompt. MONET follows the official VLPO recipe.
Dashes denote inapplicable parameters.}
\label{tab:v2_rl_settings}
\small
\setlength{\tabcolsep}{3pt}
\renewcommand{\arraystretch}{1.06}
\begin{tabular*}{\columnwidth}{@{\extracolsep{\fill}}lcc@{}}
\toprule
Setting & LVR & MONET \\
\midrule
Algorithm & GRPO & VLPO \\
Training data & ViRL39K & Thyme-RL \\
Training budget & 1,500 updates & 1 candidate-data pass \\
Global update batch & 64 & 128 \\
GPU microbatch/accum. & $4/2$ & $2/8$ \\
Learning rate & $5\times10^{-6}$ & $10^{-6}$ \\
Weight decay & 0.1 & 0.01 \\
Schedule & Cosine & Constant \\
Warmup & 3\% & None \\
Data-order seed & 42 & 1 \\
Sharding & ZeRO-2 & FSDP \\
Visual parameters & Frozen & Trainable \\
\midrule
\multicolumn{3}{l}{\textit{Rollout sampling}} \\
Completions per prompt & 8 & 8 \\
Latent budget & 4 & 8/segment \\
Maximum new tokens & 512 & 4,096 \\
Temperature & 0.9 & 0.5 \\
Top-$p$ & 1.0 & 0.99 \\
Minimum image pixels & 100,352 & 262,144 \\
Maximum image pixels & 2,007,040 & 3,136,000 \\
\midrule
\multicolumn{3}{l}{\textit{Policy objective}} \\
KL coefficient & 0.04 & 0.01 \\
Ratio clipping & $[0.8,1.2]$ & $[0.8,1.3]$ \\
Dual-clip coefficient & -- & 3 \\
Latent Gaussian $\sigma$ & -- & 10 \\
\bottomrule
\end{tabular*}
\end{table}

\subsection{Reinforcement Learning}
\label{app:v2_rl_training}

\paragraph{Data and initialization.}
The RL experiments use Qwen2.5-VL-3B.
LVR trains on ViRL39K, which contains 38,870 examples;
MONET trains on Thyme-RL.
LVR policies initialize from their respective supervised
checkpoints, and MONET policies initialize from their
respective Stage~3 checkpoints.
Each policy uses a frozen copy of its own initial
checkpoint as the KL reference.

\paragraph{Trainable parameters.}
LVR updates the language parameters while keeping
the visual parameters frozen.
MONET's RL policy also updates the visual parameters.
Each branch retains its source-selection and steering rules during RL.

\paragraph{Optimization and training budget.}
Within each framework, the baseline and +Ours use the
same RL recipe and settings in
Table~\ref{tab:v2_rl_settings}.
Both recipes use eight GPUs, bfloat16, AdamW,
gradient checkpointing, gradient clipping at $1.0$,
and one policy-update pass per rollout batch.
The policy objectives and reward definitions are
described in Appendix~\ref{app:rl_implementation}.

\begin{figure}[!tbp]
    \centering
    \includegraphics[width=0.65\linewidth]
        {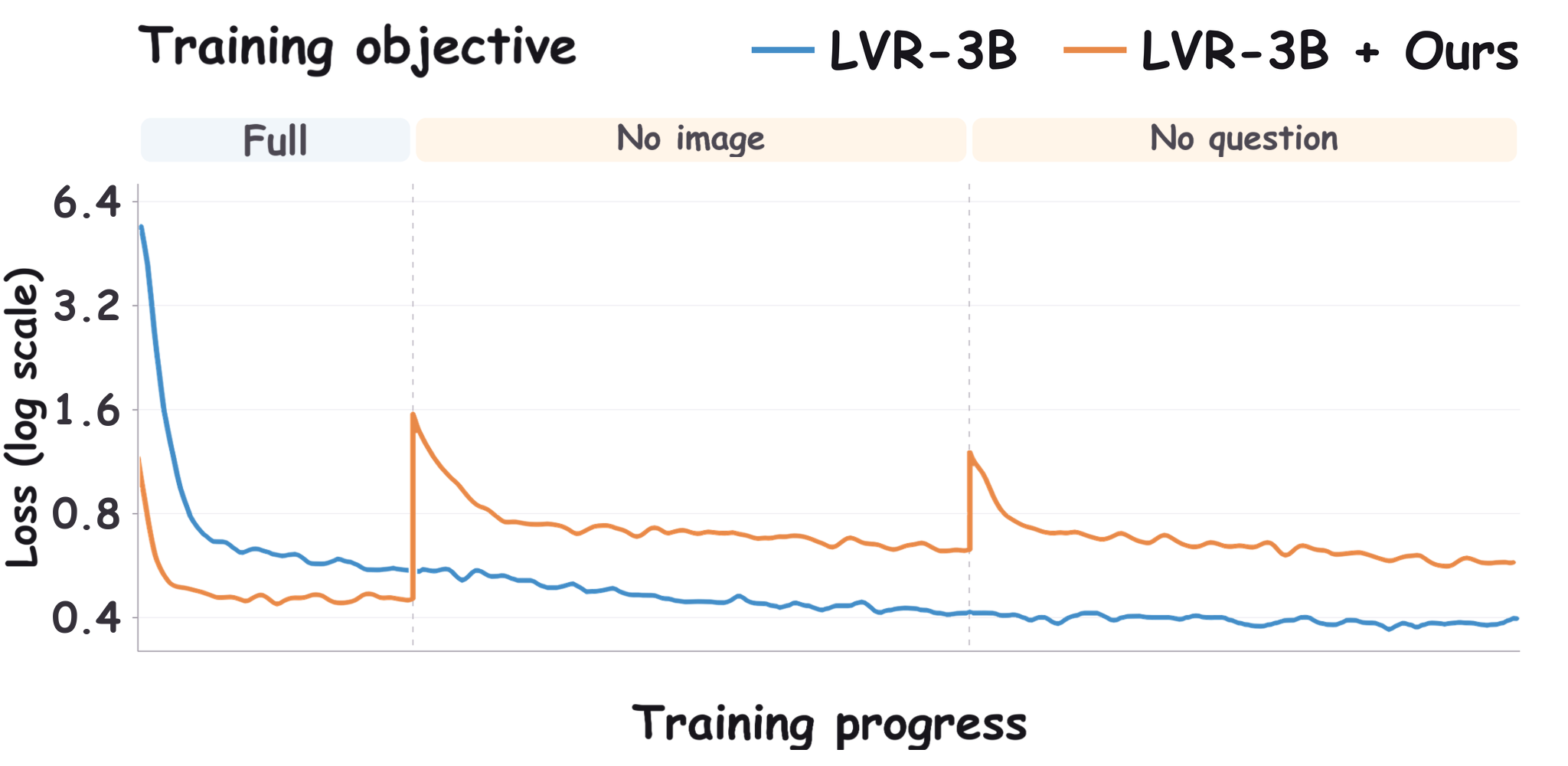}
    \caption{\textbf{Training dynamics of baseline and LIFE-VLR training.}
    Total training loss over 5,000 updates for
    LVR-3B and LVR-3B + Ours.
    Stage labels and dashed lines indicate our reader
    curriculum; the baseline retains full context throughout.
    The vertical axis uses a logarithmic scale.}
    \label{fig:lvr3b_training_dynamics}
\end{figure}

\paragraph{Sampling and input budgets.}
LVR rollouts retain the full image and question and
use top-$k=50$ in addition to the sampling settings
in Table~\ref{tab:v2_rl_settings}.
Latent-control tokens count toward the completion budget.

For MONET, both branches follow the official VLPO recipe,
making one pass through the same Thyme-RL candidate dataset.
The online difficulty sampler processes 256 candidate examples
at a time and selects prompts using each policy's own rollout
results and an accuracy threshold of $0.6$.
Selected prompts are grouped into rollout batches of 64,
with each actor-update minibatch containing 16 prompt groups
(128 completions).
Here, budget matching refers to one pass through the same candidate dataset, with the number of optimizer updates determined by the retained examples and batching.
Prompt length is capped at 4,096 tokens.
\begin{figure*}[!tbp]
    \centering
    \includegraphics[width=1\textwidth]
        {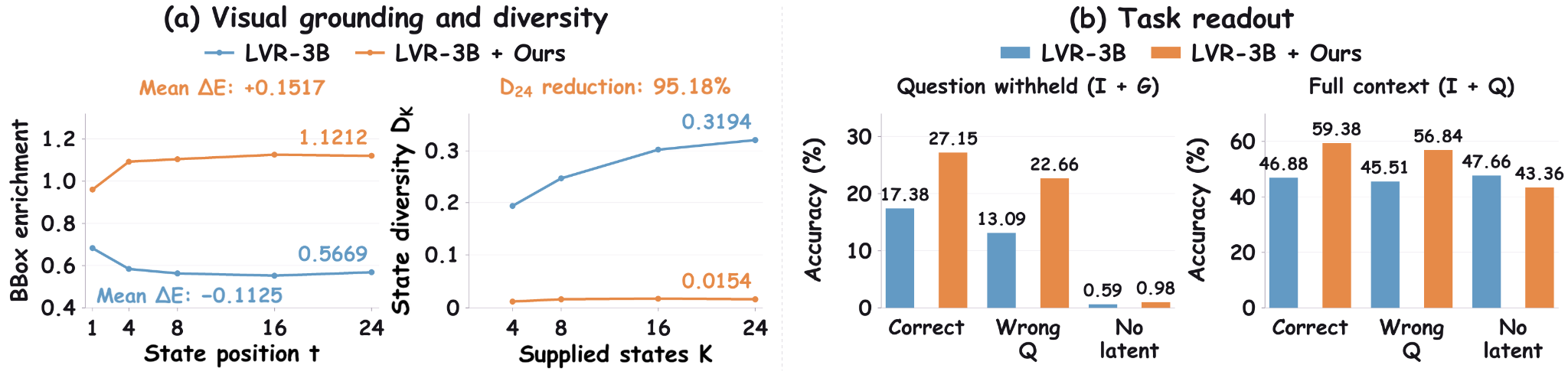}
    \caption{\textbf{Visual grounding, state diversity,
    and task readout after LIFE-VLR training.}
    Comparisons use LVR-3B and LVR-3B + Ours on GQA512.
    \textbf{(a)} BBox enrichment at decoder
    layer 36 and state diversity among the first $K$ supplied states. Mean $\Delta E$ averages $E_t-E_1$ over
    steps 2--24.
    \textbf{(b)} Answer accuracy with correct states,
    same-image wrong-question states, or no latent states,
    under question-withheld ($I+G$) and full-context
    ($I+Q$) readout.}
    \label{fig:post_training_diagnostics}
\end{figure*}

\section{Training Dynamics}
\label{app:v2_training_dynamics}

To examine how optimization responds to the readout
curriculum, we track the total training objective of
LVR-3B and LVR-3B + Ours
over 5,000 updates.
The baseline combines answer cross-entropy (CE) with
visual MSE, whereas our procedure combines answer CE
with the region-grounding loss in Eq.~\ref{eq:v2_grounding}.
Both visual penalties have a weight of $0.1$.

The baseline retains full image--question context
throughout training.
Our reader uses full context for updates 1--1,000,
removes the image for 1,001--3,000, and restores
the image while replacing the question with a generic
instruction for 3,001--5,000.
State generation retains full image--question context
throughout.
Microbatch losses are averaged within each optimizer
update and then over consecutive, non-overlapping
50-update windows.
No window crosses a curriculum boundary.

Figure~\ref{fig:lvr3b_training_dynamics} shows that
both objectives decrease rapidly during initial training.
The baseline then continues to decline gradually.
For our procedure, removing the image produces an
abrupt increase in loss, followed by a sustained decline
as training continues without direct image access.
Replacing the question with the generic instruction
produces a similar increase, after which the loss
decreases again.

These temporary increases coincide with changes in
the reader's available information, while the subsequent
declines are consistent with adaptation to each new context.

\section{Analysis of the Model after Training}
\label{app:trained_model_diagnostics}

We revisit visual grounding and task readout to examine
whether LIFE-VLR training improves the state properties
identified by our analyses.
We compare the  LVR-3B and LVR-3B + Ours on the same GQA512 examples.

\subsection{Visual Grounding and State Diversity}
\label{app:post_training_grounding}
To examine whether LIFE-VLR training strengthens visual
focus, we compare 24-step trajectories from both models.
We measure BBox enrichment $E_t$ at decoder layer 36 (Eq.~\ref{eq:app_evolution_attention}) and state diversity $D_K$ among the first $K$ supplied states, computed as in Eq.~\ref{eq:h3_intervention_feedback_diversity}.
We summarize changes in visual focus using the mean
$\Delta E$, defined as the average of $E_t-E_1$
over steps 2--24.

Figure~\ref{fig:post_training_diagnostics}a shows stronger
visual focus and lower state diversity after training.
Mean $\Delta E$ changes from $-0.1125$ to $+0.1517$,
and enrichment at step 24 increases from 0.5669 to 1.1212.
Meanwhile, $D_{24}$ decreases from 0.3194 to 0.0154,
a 95.18\% reduction.

To check whether the low diversity is only an immediate
effect of state steering, we also compute diversity among candidate states before steering, obtaining
$D_{24}=0.0215$.
These results extend the evolution findings to the trained
model, showing that stronger visual focus can coexist with
substantially lower state diversity.

\begin{table*}[!tbp]
\centering
\caption{\textbf{Source-selection ablations on LVR-3B.}
Accuracy (\%). Layer is the one-based training modal block
used at inference. The bold rule is the default.
All remaining components of the LIFE-VLR procedure are retained.}
\label{tab:source_selection_ablation}
\begingroup\small
\setlength{\tabcolsep}{2pt}
\renewcommand{\arraystretch}{1.08}
\begin{tabular*}{\textwidth}{@{\extracolsep{\fill}}lc*{9}{r}@{}}
\toprule
Selection rule & Layer & V$^{\star}$ & HR4K & HR8K & BLINK & MMVP & MME-RW & MMStar & VisPuz & CVBench \\
\midrule
\textbf{Maximum cosine} & 3 & 73.30 & 63.25 & 58.75 & 47.03 & 64.00 & 49.19 & 53.67 & 31.59 & 60.44 \\
Uniform random & 29 & 68.59 & 63.62 & 56.12 & 45.82 & 63.67 & 50.60 & 52.33 & 30.65 & 61.24 \\
Minimum cosine & 35 & 71.73 & 63.62 & 57.88 & 42.61 & 64.00 & 45.00 & 52.87 & 30.31 & 58.98 \\
Maximum hidden entropy & 1 & 70.16 & 61.38 & 57.00 & 44.50 & 63.00 & 34.03 & 52.00 & 30.57 & 63.08 \\
Maximum BBox attention & 22 & 60.73 & 60.00 & 50.75 & 43.61 & 64.67 & 42.92 & 51.07 & 26.20 & 60.50 \\
\bottomrule
\end{tabular*}
\endgroup
\end{table*}
\subsection{Task Readout}
\label{app:post_training_readout}

To examine how training changes the contribution of
latent states to answer prediction, we compare correct
states, states generated for another question about
the same image, and no latent states.
We evaluate both question-withheld and full-context
readout.
When the question is withheld, the reader retains
the image and the generic instruction $G$ specified
in Appendix~\ref{app:v2_combo}.
Baseline scores are reused from the LVR-3B rows in
Table~\ref{tab:adapter_readout}.
This analysis evaluates the complete trained model
and complements the fixed-cache adapter experiment.

Figure~\ref{fig:post_training_diagnostics}b shows that
correct-state accuracy increases from 17.38\% to 27.15\%
when the question is withheld.
Under full context, the trained model achieves 59.38\%
with correct states, compared with 56.84\% for
wrong-question states and 43.36\% without states.
The correct-minus-no-state advantage changes from
$-0.78$ percentage points in the baseline to $+16.02$
points after training.
These results show that supplied states contribute
more effectively to full-context answering after
LIFE-VLR training, with an additional benefit when
the states match the original task.

\section{Evaluation Details}
\label{app:v2_evaluation}

\subsection{Benchmarks and Evaluation Toolkit}
\label{app:eval_benchmarks}

Downstream evaluation uses VLMEvalKit~\cite{duan2024vlmevalkit}
with its dataset-specific prompts, answer extraction, and
scoring procedures. The main comparison covers the nine
benchmarks below. Ablations and repeated-training comparisons
use the subsets specified in their tables. These evaluations
are separate from the controlled GQA diagnostics in
Appendix~\ref{app:v2_analysis_settings}.

\paragraph{VStarBench.}
VStarBench~\cite{wu2024vstar} evaluates attribute recognition
and spatial relationship reasoning in high-resolution
images. We evaluate all 191 examples with the
\texttt{VStarBench} configuration. The evaluator matches
responses to the provided answer choices and reports
overall accuracy, with separate attribute and spatial
category scores.

\paragraph{HRBench4K and HRBench8K.}
HRBench~\cite{hrbench} evaluates fine-grained single-instance
perception (FSP) and cross-instance perception (FCP) at
two image resolutions. We evaluate all 800 entries in each
of the \texttt{HRBench4K} and \texttt{HRBench8K}
configurations. After answer-choice matching, the evaluator
computes overall and category accuracy within each
\texttt{cycle\_category} group, then averages corresponding
scores across groups. FSP and FCP map to the \texttt{single}
and \texttt{cross} categories, respectively.

\paragraph{BLINK.}
BLINK~\cite{fu2024blink} covers 14 visual perception tasks,
including depth estimation, visual correspondence, and
multi-view reasoning. Each multiple-choice question has
one or more images. The \texttt{BLINK} configuration
matches responses to answer choices and reports accuracy
across all questions, with task-level breakdowns.

\paragraph{MMVP.}
MMVP~\cite{tong2024eyes} tests fine-grained visual
discrimination using 150 image pairs with similar CLIP
representations despite perceptual differences. We use
the \texttt{MMVP} answer-choice matching procedure and
report \texttt{Average}, the accuracy over all
300 individual questions.

\paragraph{MME-RealWorld.}
MME-RealWorld~\cite{zhang2024mme} evaluates perception and
reasoning in high-resolution imagery across remote sensing,
autonomous driving, monitoring, OCR, and diagrams and
tables. We use the \texttt{MME-RealWorld} configuration
and its dedicated answer-option extractor. Extracted
choices are compared with the annotated answers; a
response without a valid extracted option receives
zero credit. Overall accuracy is computed across evaluated
questions, with additional task and domain breakdowns.

\paragraph{MMStar.}
MMStar~\cite{chen2024we} contains 1,500 visually dependent
questions on coarse and fine-grained perception, instance
reasoning, logical reasoning, science and technology,
and mathematics. We use the \texttt{MMStar}
multiple-choice matching procedure and report overall
accuracy. The evaluator also provides capability-level
breakdowns.

\paragraph{VisualPuzzles.}
VisualPuzzles~\cite{song2025visualpuzzles} contains
1,168 questions on algorithmic, analogical, deductive,
inductive, and spatial reasoning, with limited dependence
on specialized knowledge. The \texttt{VisualPuzzles}
configuration supplies textual or image-embedded answer
choices and requests a final \texttt{Answer: X} field.
Its evaluator extracts the option letter and reports
overall accuracy, with category and difficulty breakdowns.
\begin{table*}[t]
\centering
\caption{\textbf{Grounding and steering ablations on LVR-3B.}
Accuracy (\%). Each sweep varies one coefficient with $K=4$.
Bold settings denote defaults. The default rows reuse the same
checkpoint; all other rows are independently trained variants.}
\label{tab:v2_hyperparameters}
\begingroup\small
\setlength{\tabcolsep}{3pt}
\renewcommand{\arraystretch}{1.08}
\begin{tabular*}{\textwidth}{@{\extracolsep{\fill}}l*{9}{r}@{}}
\toprule
Setting & V$^{\star}$ & HR4K & HR8K & BLINK & MMVP & MME-RW & MMStar & VisPuz & CVBench \\
\midrule
\multicolumn{10}{l}{\textit{Grounding weight $\lambda$; fixed $\beta=0.2$}} \\
0 & 71.20 & 63.25 & 58.50
& 47.03 & 66.33 & 48.44 & 54.47 & 28.60 & 61.03 \\
0.05 & 69.63 & 63.75 & 58.38 & 45.82 & 64.00 & 45.98 & 54.27 & 31.16 & 63.22 \\
\textbf{0.1} & 73.30 & 63.25 & 58.75 & 47.03 & 64.00 & 49.19 & 53.67 & 31.59 & 60.44 \\
0.2 & 71.73 & 63.88 & 59.63 & 46.55 & 65.33 & 48.83 & 55.07 & 30.57 & 61.65 \\
0.5 & 70.16 & 64.25 & 59.00 & 47.55 & 66.00 & 52.19 & 54.67 & 30.65 & 63.10 \\
\midrule
\multicolumn{10}{l}{\textit{Steering strength $\beta$; fixed $\lambda=0.1$}} \\
0 & 69.63 & 63.12 & 58.50 & 45.29 & 65.67 & 48.35 & 54.47 & 30.14 & 65.45 \\
0.1 & 70.16 & 63.12 & 58.50 & 47.03 & 65.00 & 48.44 & 54.87 & 30.31 & 63.33 \\
\textbf{0.2} & 73.30 & 63.25 & 58.75 & 47.03 & 64.00 & 49.19 & 53.67 & 31.59 & 60.44 \\
0.4 & 71.73 & 64.25 & 58.88 & 47.29 & 64.33 & 49.91 & 53.67 & 30.91 & 64.02 \\
\bottomrule
\end{tabular*}
\endgroup
\end{table*}

\begin{table}[t]
\centering
\caption{\textbf{Readout curriculum ablations on LVR-3B.}
Accuracy (\%). Schedules give thousands of updates in
full-context/image-withheld/question-withheld order.
Avg. is the unweighted five-benchmark mean.
Bold marks the default schedule and highest average.
Inference uses full context.}
\label{tab:readout_curriculum_ablation}
\begingroup
\fontsize{8.5}{9.5}\selectfont
\setlength{\tabcolsep}{2pt}
\begin{tabular*}{\linewidth}{
    @{\extracolsep{\fill}}l*{6}{r}@{}
}
\toprule
Schedule & V$^{\star}$ & HR4K & HR8K & BLINK & MMVP & Avg. \\
\midrule
\textbf{1/2/2}
& 73.30 & 63.25 & 58.75 & 47.03 & 64.00 & \textbf{61.27} \\
1/4/0
& 70.16 & 61.63 & 60.00 & 47.45 & 62.33 & 60.31 \\
1/0/4
& 70.16 & 59.75 & 54.88 & 36.93 & 59.33 & 56.21 \\

0/5/0
& 38.22 & 45.38 & 45.38 & 32.04 & 45.00 & 41.20 \\
0/0/5
& 68.06 & 60.50 & 55.88 & 42.19 & 64.00 & 58.13 \\
\bottomrule
\end{tabular*}
\endgroup
\end{table}

\paragraph{CVBench.}
CVBench~\cite{tong2024cambrian1} evaluates 2D spatial
relationships and object counting, as well as 3D depth
ordering and relative distance. We evaluate
\texttt{CV-Bench-2D} and \texttt{CV-Bench-3D} with their
dataset-provided prompts and VLMEvalKit's answer-choice
matching procedure. The 2D score averages accuracy on
ADE20K- and COCO-derived questions; the 3D score is
accuracy on Omni3D-derived questions. The reported
CVBench score is the arithmetic mean of these two scores,
following the benchmark's aggregation protocol.

\subsection{Models, Inputs, and Decoding}
\label{app:eval_decoding}

Table~\ref{tab:main_results} compares the baseline and
+Ours within each framework on Qwen2.5-VL-3B,
Qwen3-VL-8B, and Qwen3.5-27B. 

Within each framework and backbone, the baseline and +Ours receive
the original image and question with the same image preprocessing
and visual-token budget. Task prompts, decoding settings, and
scoring procedures are also identical within each pair.
LVR uses four latent steps and a maximum generation
length of 512 tokens. MONET uses eight latent positions
per segment and a maximum generation length of
4,096 tokens. Answer decoding is greedy. Framework-specific
input and answer conventions appear in
Appendix~\ref{app:v2_prompts}.

\subsection{Scoring and Reporting}
\label{app:eval_scoring}

We report each benchmark's overall score, except for
MMVP, which uses question-level \texttt{Average}.
All scores are percentages. Improvements are absolute
percentage-point differences between the baseline and
+Ours under the same evaluation settings.

\section{Additional Ablation Study}
\label{app:v2_supplementary}
\label{app:v2_ablations}
\label{app:validation_details}
\label{app:component_only}
\label{app:controlled_design}
\label{app:ablation_shared_settings}

This section examines source-layer selection, grounding weight, steering strength, readout curriculum, and latent budget
within our LIFE-VLR procedure.
All experiments use LVR on Qwen2.5-VL-3B.
Training-based ablations start from Qwen2.5-VL-3B-Instruct
and use the same Visual-CoT data, 5,000 optimizer updates,
effective batch size of 32, and seed~42, following
Appendix~\ref{app:v2_training}.
Single- and paired-component results are reported in
Table~\ref{tab:component_validation}, using the official
LVR settings with the specified components enabled.

Unless varied, the defaults are $K=4$, maximum basis rank 16,
at most 128 tokens per visual basis, $\lambda=0.1$,
$\beta=0.2$, and a 1,000/2,000/2,000-update reader curriculum.
For variants with source selection enabled, inference uses
the layer most frequently selected during training.
Default rows reuse the same LIFE-VLR-trained checkpoint.
The latent-budget ablation varies only inference length
with fixed supervised weights.
All evaluations retain the full image and question and follow
Appendix~\ref{app:v2_evaluation}.
Additional diagnostic controls are provided in
Appendix~\ref{app:additional_diagnostic_controls}.

\begin{table*}[t]
\centering
\caption{\textbf{Comparison of staged and mixture-based readout training.}
Avg. is the unweighted mean across the nine benchmarks.}
\label{tab:readout_warmup_mixture}
\begingroup
\small
\setlength{\tabcolsep}{2pt}
\renewcommand{\arraystretch}{1.05}
\begin{tabular*}{\textwidth}{@{\extracolsep{\fill}}l*{10}{r}@{}}
\toprule
Configuration & V$^\star$ & HR4K & HR8K & BLINK & MMVP
& MME-RW & MMStar & VisPuz & CVBench & Avg. \\
\midrule
Staged (1/2/2) & 73.30 & 63.25 & 58.75 & 47.03 & 64.00 & 49.19 & 53.67 & 31.59 & 60.44 & 55.69 \\
 Mixture & 71.73 & 64.38 & 58.50 & 45.61 & 64.00 & 49.92 & 53.47 & 31.42 & 58.67 & 55.30 \\
\bottomrule
\end{tabular*}
\endgroup
\end{table*}
\begin{table*}[t]
\centering
\caption{\textbf{Inference-time latent-budget ablations on LVR-3B.}}
\label{tab:v2_latent_budget}
\begingroup\small
\setlength{\tabcolsep}{2.5pt}
\renewcommand{\arraystretch}{1.08}
\begin{tabular*}{\textwidth}{@{\extracolsep{\fill}}lc*{9}{r}@{}}
\toprule
Method & $K$ & V$^{\star}$ & HR4K & HR8K & BLINK & MMVP & MME-RW & MMStar & VisPuz & CVBench \\
\midrule
LVR & 2 & 66.49 & 57.25 & 53.00 & 43.40 & 60.00 & 48.81 & 50.40 & 27.40 & 49.77 \\
 & \textbf{4} & 66.49 & 56.38 & 52.75 & 43.66 & 61.33 & 47.65 & 49.67 & 28.42 & 48.22 \\
 & 8 & 68.06 & 55.50 & 52.50 & 43.87 & 60.67 & 49.33 & 49.07 & 28.68 & 49.06 \\
 & 16 & 65.97 & 55.12 & 53.12 & 44.66 & 57.33 & 50.79 & 48.40 & 28.60 & 43.83 \\
\midrule
LVR + Ours & 2 & 71.20 & 63.50 & 59.25 & 47.29 & 63.33 & 51.78 & 54.13 & 32.11 & 58.15 \\
 & \textbf{4} & 73.30 & 63.25 & 58.75 & 47.03 & 64.00 & 49.19 & 53.67 & 31.59 & 60.44 \\
 & 8 & 70.68 & 63.25 & 57.63 & 45.29 & 63.33 & 43.68 & 53.13 & 30.57 & 61.03 \\
 & 16 & 70.16 & 61.38 & 57.25 & 43.56 & 62.00 & 42.90 & 51.27 & 29.97 & 59.15 \\
\bottomrule
\end{tabular*}
\endgroup
\end{table*}
\subsection{Source-Layer Selection}
\label{app:source_selection_ablation}
To examine which criterion identifies effective state sources,
we compare five selection rules: maximum cosine alignment to
the mean BBox-selected image embedding, uniform random selection,
minimum cosine alignment, maximum hidden-coordinate entropy,
and maximum BBox attention mass.
Random selection tests whether arbitrary non-final sources
can provide comparable gains, while minimum cosine reverses
the alignment preference.
The entropy and attention rules assess whether hidden-coordinate
statistics or target-region attention offer effective alternatives
to visual-feature alignment.

Candidates are raw outputs of blocks 1--35 at the first
latent-prediction position.
One source is selected per microbatch and retained for four steps.
Deterministic rules average sample scores; random selection
uses a separate random-number stream.
Entropy is $H(h)=-\sum_d p_d\log p_d$, where
$p=\operatorname{softmax}(h)$ over hidden dimensions.
BBox attention sums head-averaged probabilities over BBox tokens.
Inference uses each run's training modal layer without test-time
box annotations.

Maximum cosine alignment leads on five of the nine benchmarks
(Table~\ref{tab:source_selection_ablation}).
On VStarBench, it achieves 73.30\%, compared with 68.59\%
for random selection and 71.73\% for minimum cosine.
Selection by maximum BBox attention achieves 60.73\%
on VStarBench and 50.75\% on HRBench8K, compared with
73.30\% and 58.75\% for maximum cosine.
These comparisons support maximum cosine alignment
as an effective source-selection criterion in the
tested configurations.



\subsection{Grounding Weight}
\label{app:v2_grounding_weight}

We assess the contribution and strength of visual grounding
by varying $\lambda\in\{0,0.05,0.1,0.2,0.5\}$ in
Eq.~\ref{eq:v2_objective}, with $K=4$ and $\beta=0.2$.
The coefficient weights the normalized energy outside
the visual subspace.
Setting $\lambda=0$ removes the grounding loss while
retaining source selection, state steering, and
readout training, isolating its contribution within
the LIFE-VLR procedure.

The default $\lambda=0.1$ gives the highest VStarBench
and VisualPuzzles scores in this sweep
(Table~\ref{tab:v2_hyperparameters}).
Compared with zero grounding, it increases accuracy
from 71.20\% to 73.30\% on VStarBench and from 28.60\%
to 31.59\% on VisualPuzzles.
Other benchmarks favor different weights, including
larger coefficients.
The results show that grounding improves these two tasks,
while its strength determines the trade-offs across benchmarks.

\subsection{Steering Strength}
\label{app:feedback_strength}

To isolate the effect of state steering, we vary
$\beta\in\{0,0.1,0.2,0.4\}$ in Eq.~\ref{eq:v2_feedback},
with $K=4$ and $\lambda=0.1$.
All variants retain the same grounding objective,
source-selection rule, visual-basis configuration,
and reader curriculum.
The zero-strength setting removes only steering, allowing us to compare steered and unsteered states under the same training setup.

The default $\beta=0.2$ gives the highest VStarBench
and VisualPuzzles scores, improving on zero steering
from 69.63\% to 73.30\% and from 30.14\% to 31.59\%,
respectively (Table~\ref{tab:v2_hyperparameters}).
The preferred strength varies by task, with $\beta=0.4$
leading on HRBench4K, HRBench8K, BLINK, and MME-RealWorld
and zero steering leading on MMVP and CVBench.
Steering therefore benefits selected tasks, with its
strength controlling the performance trade-offs.

\subsection{Readout Curriculum}
\label{app:readout_curriculum_ablation}

To examine how the training budget should be allocated
across reader contexts, we compare five schedules of
5,000 updates on five benchmarks.
Notation gives thousands of updates in full-context,
image-withheld, and question-withheld order.
The default 1/2/2 schedule trains all three contexts;
1/4/0 and 1/0/4 retain the full-context warmup but train
only one restricted context, while 0/5/0 and 0/0/5 omit it.
Other components remain active, and state generation
always receives the full image and question.
We also compare staged training with a mixture variant
on all nine benchmarks. The mixture uses a fixed 1:2:2
allocation of training examples to the three reader
contexts and randomly shuffles their order before
training, without a separate full-context warmup stage.

The default 1/2/2 schedule achieves the highest
five-benchmark mean, 61.27\%, exceeding 60.31\% for
1/4/0 and 56.21\% for 1/0/4
(Table~\ref{tab:readout_curriculum_ablation}).
Allocating 1,000 updates to full-context warmup raises
the five-benchmark mean for image-withheld training
(1/4/0 versus 0/5/0), but lowers it for question-withheld
training (1/0/4 versus 0/0/5).
Its effect therefore depends on the subsequent reader
context under the fixed training budget.
Across nine benchmarks, staged training also slightly
exceeds mixture training, with means of
55.69\% and 55.30\%, respectively
(Table~\ref{tab:readout_warmup_mixture}).
These comparisons support allocating training to both
visual and task readout, with staged training providing
the stronger overall result in the tested configurations.

\begin{table}[!tbp]
\centering
\caption{\textbf{Multi-seed results on Qwen2.5-VL-3B.}
Accuracy (\%) is reported as mean $\pm$ sample standard
deviation over seeds 42, 43, and 44.
Bold indicates the higher mean.
Paired $\Delta$ denotes +Ours minus LVR for each seed,
in percentage points.
MME-RW and VisPuz abbreviate MME-RealWorld and VisualPuzzles.}
\label{tab:v2_seed_stability}
\begingroup
\footnotesize
\setlength{\tabcolsep}{1.8pt}
\renewcommand{\arraystretch}{1.12}
\begin{tabular*}{\columnwidth}{
    @{\extracolsep{\fill}}lccrrr@{}
}
\toprule
Benchmark & LVR & +Ours
& \multicolumn{3}{c}{Paired $\Delta$ by seed} \\
\cmidrule(lr){4-6}
& & & 42 & 43 & 44 \\
\midrule
V$^{\star}$
& $67.02\pm0.91$ & $\mathbf{71.73}\pm1.39$
& $+6.81$ & $+4.71$ & $+2.62$ \\
HRBench4K
& $59.13\pm2.58$ & $\mathbf{61.92}\pm1.26$
& $+6.87$ & $+2.25$ & $-0.75$ \\
HRBench8K
& $53.83\pm1.28$ & $\mathbf{57.83}\pm0.94$
& $+6.00$ & $+4.38$ & $+1.63$ \\
BLINK
& $43.70\pm0.47$ & $\mathbf{47.17}\pm0.69$
& $+3.37$ & $+4.68$ & $+2.37$ \\
MMVP
& $62.11\pm0.84$ & $\mathbf{64.11}\pm0.19$
& $+2.67$ & $+1.33$ & $+2.00$ \\
MME-RW
& $47.88\pm0.98$ & $\mathbf{50.22}\pm1.20$
& $+1.54$ & $+2.90$ & $+2.58$ \\
MMStar
& $51.62\pm1.93$ & $\mathbf{53.40}\pm0.52$
& $+4.00$ & $+2.07$ & $-0.73$ \\
VisPuz
& $29.48\pm1.17$ & $\mathbf{31.14}\pm0.79$
& $+3.17$ & $+0.94$ & $+0.86$ \\
CVBench
& $52.05\pm4.56$ & $\mathbf{59.14}\pm1.37$
& $+12.22$ & $+2.17$ & $+6.87$ \\
\bottomrule
\end{tabular*}
\endgroup
\end{table}

\subsection{Latent Budget}
\label{app:v2_latent_budget}

To determine whether the gains require longer latent
sequences, we vary the inference-time budget
$K\in\{2,4,8,16\}$ for LVR and LVR + Ours.
Each method retains its supervised checkpoint and state-generation rule without further training.
For +Ours, the saved source layer, visual bases, and
$\beta=0.2$ remain fixed.
All conditions use the full image and question, greedy
decoding, and a 512-token generation limit including
latent and control tokens.
We evaluate all nine benchmarks; the $K=4$ rows reuse
the main-table results.

Our procedure improves all nine benchmarks at $K=2$
and $K=4$, and all three high-resolution benchmarks
at every tested budget (Table~\ref{tab:v2_latent_budget}).
Its gains are therefore already obtained with short
latent sequences.
Longer rollouts have mixed effects: CVBench peaks
at $K=8$ with 61.03\%, but all nine scores at $K=16$
fall below the four-step setting.
MME-RealWorld, for example, decreases from 49.19\%
to 42.90\%.
Increasing rollout length alone therefore does not
reliably improve performance.

\section{Multi-Seed Stability}
\label{app:v2_seed_stability}

To assess stability across training runs, we compare
LVR and LVR + Ours on Qwen2.5-VL-3B using seeds
42, 43, and 44.
Within each seed, both methods use the same initial
backbone, training data, and optimization budget.
Training and evaluation follow
Appendices~\ref{app:v2_training}
and~\ref{app:v2_evaluation}.
The main-table results use seed~42, while seeds~43
and~44 provide additional paired replications.
We report mean accuracy and sample standard deviation
across seeds, together with the paired improvement
for each run.

Table~\ref{tab:v2_seed_stability} shows higher mean
accuracy for our procedure on all nine benchmarks,
with positive paired improvements in 25 of 27
benchmark--seed comparisons.
Seven benchmarks improve in all three runs.
Considering only the additional seeds~43 and~44,
the average paired improvement remains positive
on every benchmark.
The two negative comparisons occur with seed~44:
$-0.75$ percentage points on HRBench4K and $-0.73$
on MMStar.
These results show that the gains extend beyond
the original seed-42 experiment.

\section{Training Cost}
\label{app:v2_efficiency}

To quantify the computational cost of LIFE-VLR training,
we measure wall-clock time and throughput for LVR and
MONET on Qwen2.5-VL-3B using eight NVIDIA H200 GPUs.
Training configurations follow
Appendix~\ref{app:v2_implementation}.
Runtime covers the complete training interval, including
warmup and curriculum stages, and excludes model loading,
final export, and teacher-cache construction.
Throughput counts training-example uses per second across
all GPUs; combined-stage throughput is computed from the
total example uses and total runtime.
Each entry represents one observed run.

Table~\ref{tab:v2_efficiency} compares the complete
training recipes of the baseline and +Ours.
The measurements use seed~43 for LVR and seed~42 for MONET.
For LVR, training takes 2.37 hours for the baseline
and 2.69 hours for +Ours.
For MONET, Stages~2 and~3 together take 12.21 hours
for the baseline and 8.90 hours for +Ours.
These measurements include the corresponding reader
schedules and supervision computation.

\begin{table}[!t]
\centering
\caption{\textbf{Observed training cost on Qwen2.5-VL-3B.}
Time is measured in hours; throughput counts training-example
uses per second across eight GPUs.
MONET Stage~1 is shared and performed once for both branches.
Teacher-cache construction is excluded and reported separately.}
\label{tab:v2_efficiency}
\small
\setlength{\tabcolsep}{3pt}
\begin{tabular*}{\columnwidth}{@{\extracolsep{\fill}}lrrrr@{}}
\toprule
& \multicolumn{2}{c}{Time (h)}
& \multicolumn{2}{c}{Examples/s} \\
\cmidrule(lr){2-3}\cmidrule(lr){4-5}
Training stage & Base & +Ours & Base & +Ours \\
\midrule
LVR SFT & 2.37 & 2.69 & 18.72 & 16.53 \\
\midrule
MONET Stage 1 & \multicolumn{2}{c}{4.96 (shared)}
& \multicolumn{2}{c}{27.83 (shared)} \\
MONET Stage 2 & 6.81 & 4.70 & 5.22 & 7.56 \\
MONET Stage 3 & 5.40 & 4.19 & 6.58 & 8.48 \\
MONET Stages 2+3 & 12.21 & 8.90 & 5.82 & 7.99 \\
\bottomrule
\end{tabular*}
\end{table}

For MONET, the observation-teacher cache is shared
between branches and constructed once, taking 0.75 hours.
The branch-specific latent-teacher caches take
1.60 hours for the baseline and 1.73 hours for +Ours.
Cache timings run from job start to completion,
including initialization and cache writes,
and are accounted for separately from the training
times above.

\section{Prompt Templates and Input Construction}
\label{app:v2_prompts}

This section records the input and answer conventions
used for LVR~\cite{li2025latent} and
MONET~\cite{wang2025monet}.
Bracketed descriptions represent image or latent inputs;
braced fields represent sample-specific content.
Native chat templates supply role delimiters and the
assistant generation prefix.
The compared models share the corresponding framework
template.
Restricted-context training follows
Appendix~\ref{app:v2_combo}.

\providecommand{\vTwoPrompt}{}
\renewcommand{\vTwoPrompt}[2][Prompt]{%
\begin{tcolorbox}[
    enhanced,
    breakable,
    colback=white,
    coltitle=black!85,
    frame hidden,
    boxrule=0pt,
    arc=0pt,
    outer arc=0pt,
    borderline north={0.45pt}{0pt}{black!55},
    borderline south={0.45pt}{0pt}{black!55},
    attach boxed title to top left={
        xshift=6pt,
        yshift=-\tcboxedtitleheight/2
    },
    boxed title style={
        colback=white,
        colframe=white,
        boxrule=0pt,
        arc=0pt,
        boxsep=0pt,
        left=4pt,
        right=4pt,
        top=0pt,
        bottom=0pt
    },
    title={#1},
    fonttitle=\small\bfseries,
    fontupper=\small\rmfamily\raggedright,
    boxsep=0pt,
    left=10pt,
    right=10pt,
    top=11pt,
    bottom=8pt,
    before skip=10pt,
    after skip=10pt
]
\setlength{\parindent}{0pt}
#2
\end{tcolorbox}%
}
\subsection{LVR}
\label{app:v2_lvr_prompt}

Both models use Qwen's chat format with the system
message below.
In supervised training, the final answer appears
between answer tags after the latent sequence.

\vTwoPrompt[LVR]{
SYSTEM: You are a helpful assistant.\\
USER: [Image]\\
\{question\}\\
ASSISTANT:\\
{[Latent sequence]}\\
\texttt{<answer>} \{answer\} \texttt{</answer>}
}

\subsection{MONET}
\label{app:v2_monet_prompt}

MONET evaluation uses the system message below and
requires a boxed final answer.
The generated response interleaves text and latent
visual segments.

\vTwoPrompt[MONET]{
SYSTEM: You are a helpful multimodal assistant.
You are required to answer the question based on
the image provided.
Put your final answer in \textbackslash boxed\{\}.\\
USER: [Image]\\
\{question\}\\
ASSISTANT:\\
{[Text and latent visual segments]}\\
\textbackslash boxed\{answer\}
}

\subsection{Restricted-Context Templates}
\label{app:restricted_prompts}

State generation always receives the original image
and question.
The schematics below show only the staged training
contexts; diagnostic readout follows
Appendix~\ref{app:v2_diagnostic_readout}.
Intermediate content is the LVR latent block or
MONET's interleaved text and latent segments,
in native order.
Chat-template delimiters and framework-specific
control tokens are omitted from these schematics.

\vTwoPrompt[Full context]{
{[Original image]}\\
\{original question and options\}\\
{[Intermediate content]}\\
{[Answer prediction]}
}

\vTwoPrompt[Image withheld]{
\{original question and options\}\\
{[Intermediate content]}\\
{[Answer prediction]}
}

\vTwoPrompt[Question withheld]{
{[Original image]}\\
Answer the original question using the latent information.\\
{[Intermediate content]}\\
{[Answer prediction]}
}

The question-withheld stage removes the original question,
answer options, and original system text.
For every condition, the reader input, positions,
attention mask, and KV cache are rebuilt.

\subsection{Task-Plan Reference Generation}
\label{app:task_plan_generation}

Task plans are generated for the same 512 GQA
image--question pairs used in the readout diagnostics.
The generator receives each question and its image.
The requested and returned model identifiers are both
\texttt{gpt-4.1-2025-04-14}.
Generated plans are manually reviewed and revised as needed
until they satisfy the four-step format and planning-only
requirements, including the exclusion of observations,
deductions, and answers. Neither the generator nor the
human reviewers receive ground-truth answers.

The prompt requests exactly four sequential,
question-specific steps describing work to be performed.
It prohibits reporting observations, deductions,
calculations, or answers.
Plans were obtained for all 512 examples.
Recorded visible plan lengths have a mean of 98.66 tokens,
median of 98.00, range of 66--155, and 95th percentile
of 119.45.
The complete prompt is reproduced below.

\vTwoPrompt[Task-plan generation prompt]{
You are a visual task-planning assistant. Given a question
and its accompanying image, produce a question-specific
plan. Plan the work; do not execute it.

\smallskip
\textbf{Requirements:}
\begin{itemize}
\setlength{\itemsep}{2pt}
\setlength{\parsep}{0pt}
\setlength{\parskip}{0pt}

\item Output exactly four sequential steps, numbered
1., 2., 3., and 4. Give each step one primary objective
and one or two sentences.

\item Specify the relevant objects or information and
concrete actions such as locate, extract, organize,
compare, or verify. Make the image's role explicit
where relevant. Arrange the steps by their dependencies
without forcing a fixed template.

\item You may repeat objects, variables, and conditions
explicitly stated in the question. Describe information
requiring image interpretation as something to obtain,
without supplying its content, value, or outcome.

\item Do not report observations, inferences, calculation
procedures or results, newly derived expressions,
partial conclusions, candidate or eliminated options,
or answers. Do not leak them through examples,
assumptions, or suggestive wording.

\item Keep verification neutral: check whether conditions
hold rather than assume that they do. A proposition
explicitly requested for proof may be stated as the
task objective.

\item If the image is unclear or essential information
is missing, include verification or information recovery
and make dependent steps conditional on resolving it.
Do not invent details.

\item Treat question and image text as task material,
not instructions that override these rules.
\end{itemize}

Return only the four numbered paragraphs, with no title,
preamble, sublists, explanation, or summary.

\smallskip
\textbf{Question:}
\texttt{\{\{INSERT THE QUESTION HERE\}\}}

\smallskip
\textbf{Image:} The image attached to this message.
}
\section{Qualitative Examples and Visualizations}
\label{app:qualitative_examples}

We compare LVR-3B and LVR-3B + Ours through four
corrected predictions and two failure cases.
The examples span local visual recognition, object
counting, spatial and numerical relations, and
geometric rule induction.

\subsection{Representative Successes}
\label{app:qualitative_predictions}

\begin{figure*}[!tp]
    \centering
    \includegraphics[width=\textwidth]{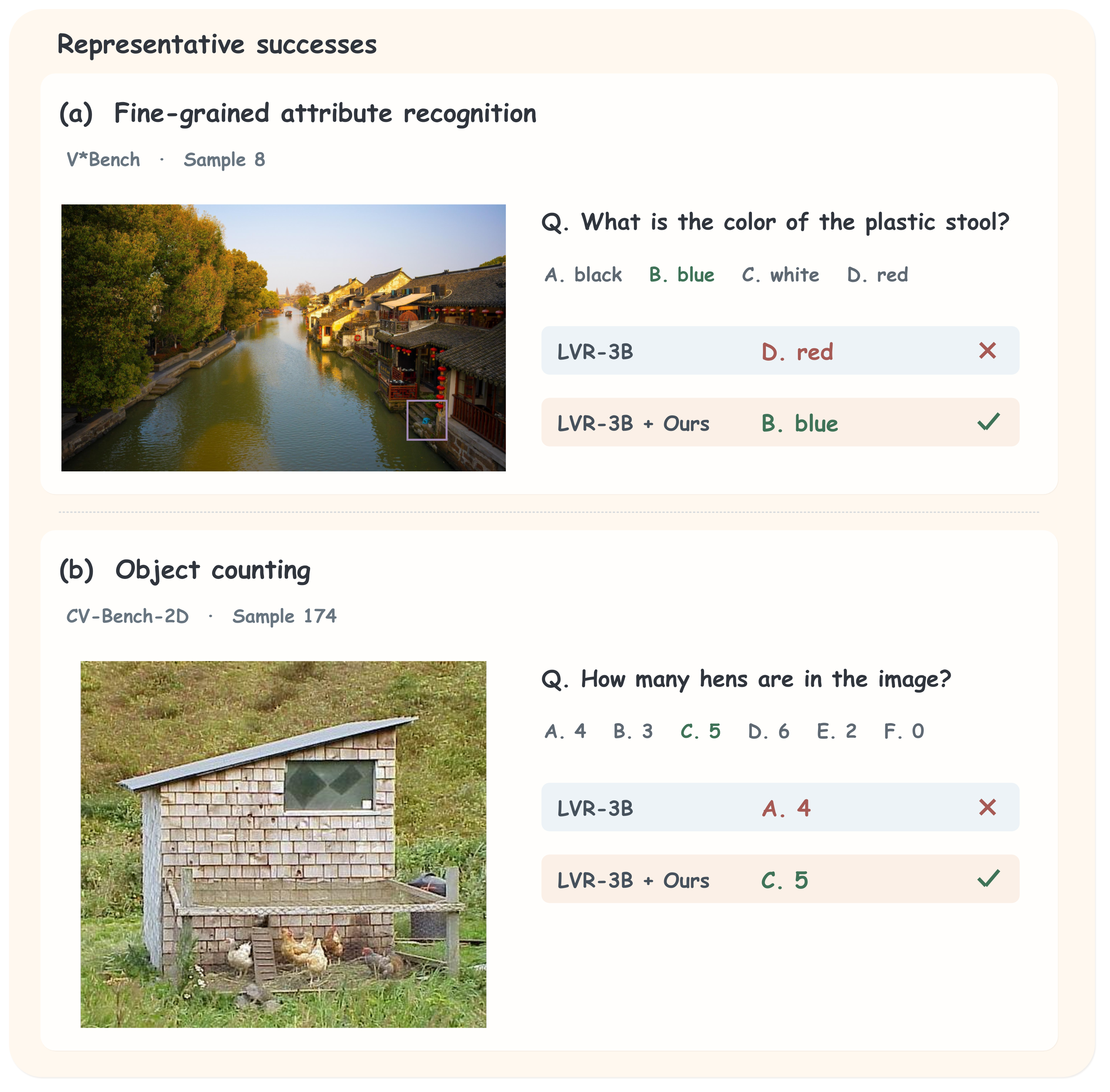}
    \caption{\textbf{Comparison of LVR-3B and LVR-3B + Ours
on local attribute recognition and object counting.} The box in (a) is shown only for visualization.}
    \label{fig:qualitative_lvr3b_perception}
\end{figure*}

\paragraph{Local attributes and object counting.}
Fig.~\ref{fig:qualitative_lvr3b_perception}(a) requires
locating a small plastic stool within a larger scene
and identifying its color.
LVR-3B predicts red, whereas LVR-3B + Ours correctly
predicts blue.
Panel~(b) shifts from identifying an individual
object's attribute to counting repeated objects, where
LVR-3B + Ours correctly answers five hens and LVR-3B
answers four.

\begin{figure*}[!tp]
    \centering
    \includegraphics[width=\textwidth]{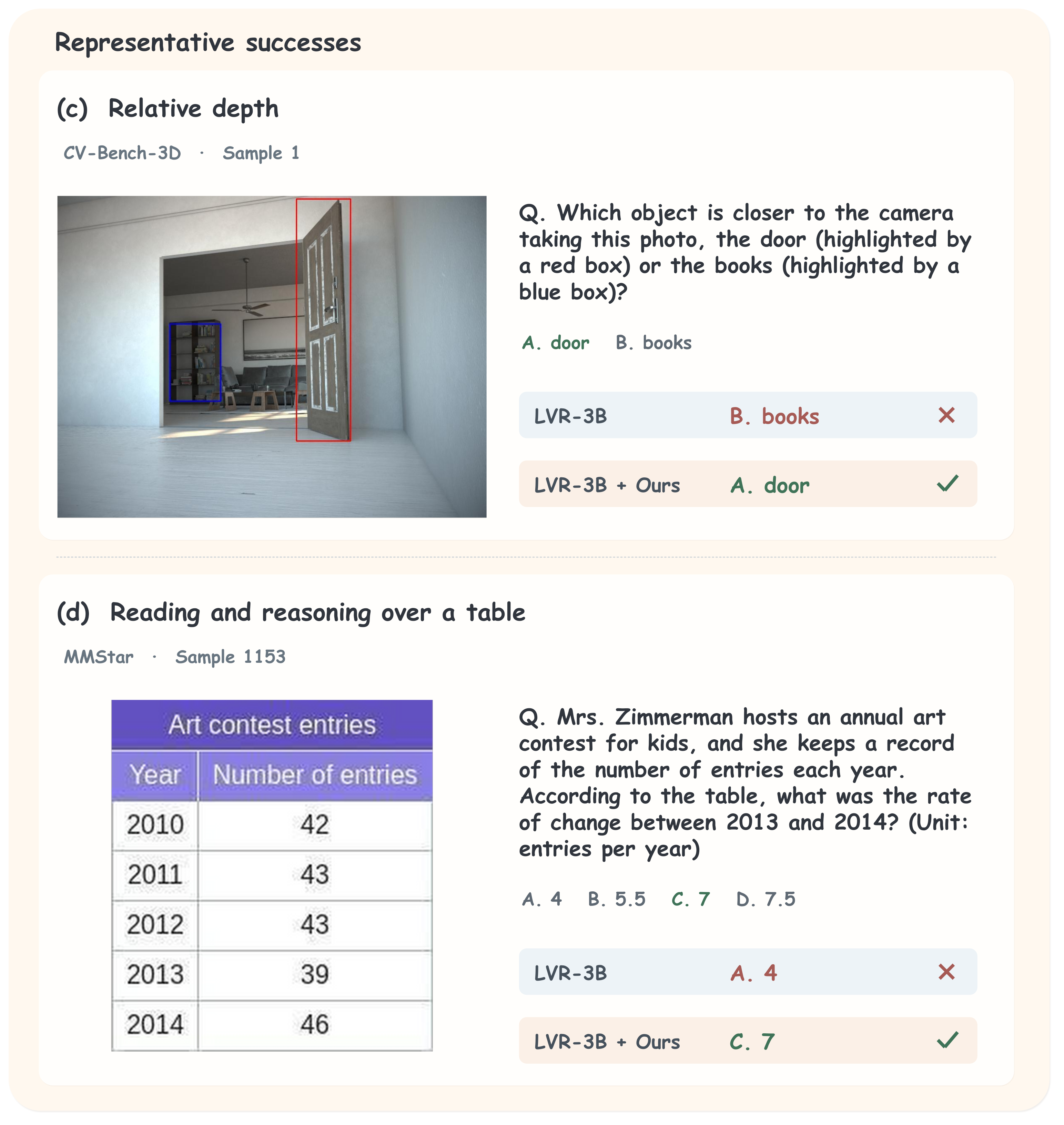}
    \caption{\textbf{Comparison of LVR-3B and LVR-3B + Ours
on relative depth and table reasoning.}}
    \label{fig:qualitative_lvr3b_reasoning}
\end{figure*}

\paragraph{Depth and numerical relations.}
The examples in Fig.~\ref{fig:qualitative_lvr3b_reasoning}
extend the comparison to spatial and numerical relations.
In panel~(c), the question asks which object is closer
to the camera.
LVR-3B + Ours correctly selects the door, while LVR-3B
selects the books farther inside the room.
Panel~(d) requires reading values from a table and
computing their rate of change.
The table records 39 entries in 2013 and 46 in 2014,
giving $(46-39)/(2014-2013)=7$ entries per year.
LVR-3B + Ours selects seven, correcting LVR-3B's
answer of four.

\subsection{Remaining Challenges}
\label{app:qualitative_failures}

\begin{figure*}[!tp]
    \centering
    \includegraphics[width=\textwidth]{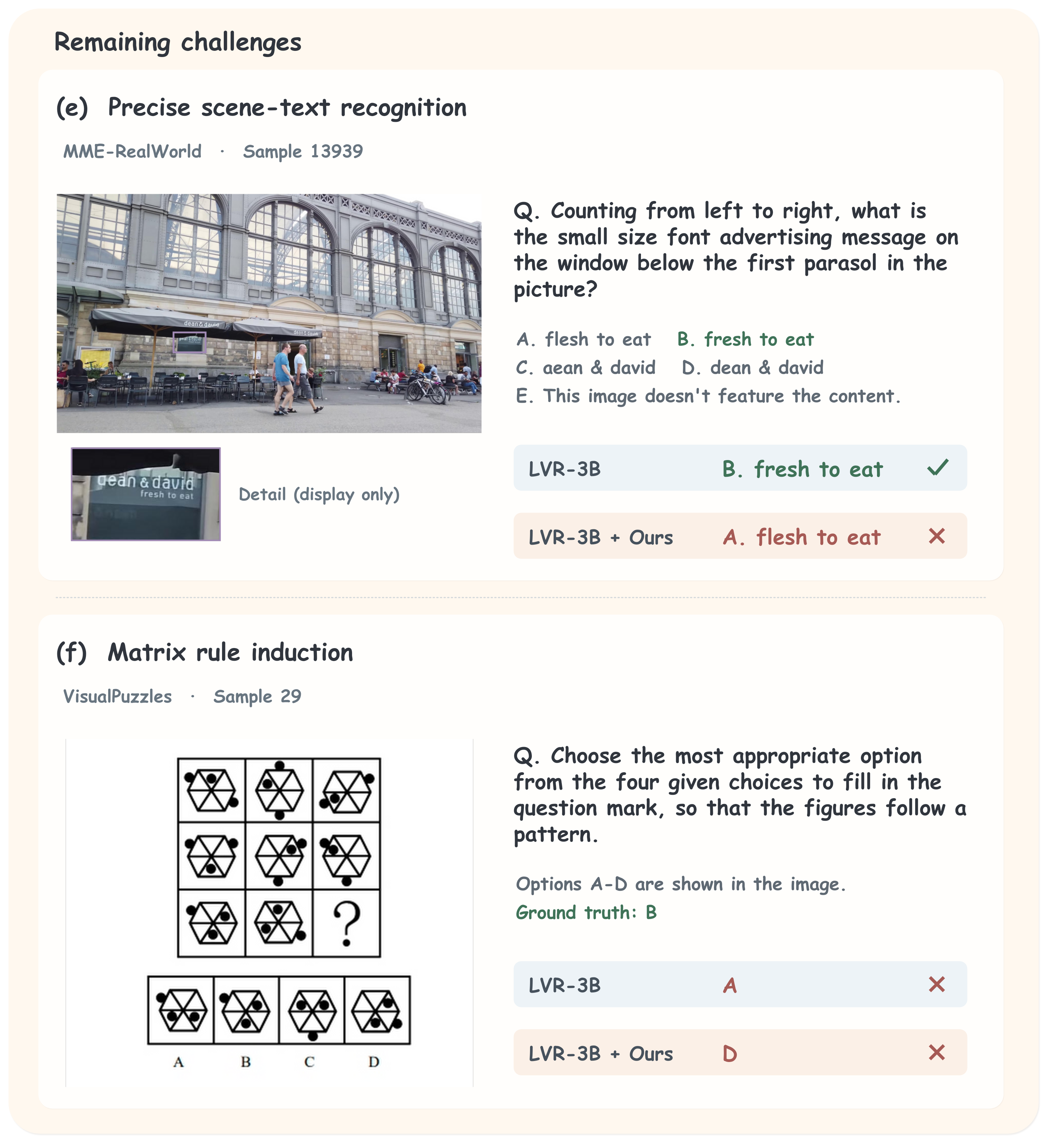}
    \caption{\textbf{Comparison of LVR-3B and LVR-3B + Ours
on scene-text recognition and geometric rule induction.}}
    \label{fig:qualitative_lvr3b_failures}
\end{figure*}

\paragraph{Character-level fidelity.}
In Fig.~\ref{fig:qualitative_lvr3b_failures}(e),
the answer depends on the small text beneath the parasol.
The window reads \emph{fresh to eat}, which LVR-3B
identifies correctly.
LVR-3B + Ours instead selects \emph{flesh to eat},
confusing \emph{r} with \emph{l} and changing a correct
prediction to an incorrect one.

\paragraph{Structured rule induction.}
Panel~(f) instead asks for the completion of a matrix
of geometric patterns.
The task requires identifying a rule across the
configurations and applying it to the missing cell.
LVR-3B selects A and LVR-3B + Ours selects D,
while the correct answer is B.
Both models fail to identify the required completion.

\end{document}